\documentclass{article}

\PassOptionsToPackage{numbers, compress}{natbib}

\usepackage[main,final]{neurips_2026}

\usepackage{graphicx}
\usepackage{algorithm}
\usepackage{algpseudocode}
\usepackage{pifont}
\usepackage{enumitem}

\usepackage[utf8]{inputenc} % allow utf-8 input
\usepackage[T1]{fontenc}    % use 8-bit T1 fonts
\usepackage{hyperref}       % hyperlinks
\usepackage{url}            % simple URL typesetting
\usepackage{booktabs}       % professional-quality tables
\usepackage{amsfonts}       % blackboard math symbols
\usepackage{nicefrac}       % compact symbols for 1/2, etc.
\usepackage{microtype}      % microtypography
\usepackage{xcolor}         % colors
\usepackage{nicematrix}
\usepackage[normalem]{ulem}

\usepackage{amsmath,amsfonts,bm,amssymb}

\def\eqref#1{Eq.~\ref{#1}}
\def\1{\bm{1}}

\DeclareMathAlphabet{\mathsfit}{\encodingdefault}{\sfdefault}{m}{sl}
\SetMathAlphabet{\mathsfit}{bold}{\encodingdefault}{\sfdefault}{bx}{n}

\newcommand{\KS}{K(\mathcal{S})}

\newcommand{\bigO}{\mathcal{O}}
\newcommand{\drop}[1]{{\scriptsize\textcolor{red!70!black}{($#1$)}}}

\title{Retrieval from Within:\\ An Intrinsic Capability of Attention-Based Models}

\author{
Elad Hoffer$^1$,\,
Yochai Blau$^1$,\,
Edan Kinderman$^1$,
Ron Banner$^1$,\,
\\[0.2cm]
\textbf{Daniel Soudry$^{1,2}$,\,
Boris Ginsburg$^1$}
\\[0.2cm]
$^1$NVIDIA \quad $^2$Technion
\\[0.2cm]
\small{\texttt{\{elad.hoffer,daniel.soudry\}@gmail.com}}\\
\small{\texttt{\{{yblau},{ekinderman},{rbanner},{bginsburg}\}@nvidia.com}}
}

\begin{document}

\maketitle

\begin{abstract}

Retrieval-augmented generation (RAG) typically treats retrieval and generation as separate systems.
We ask whether an attention-based encoder-decoder can instead retrieve directly from its own internal representations.
We introduce \emph{INTRA} (\emph{INTrinsic Retrieval via Attention}), 
a framework where decoder attention queries score pre-encoded evidence chunks that are then directly reused as context for generation.
By construction, INTRA unifies retrieval and generation, eliminating the retriever-generator mismatch typical of RAG pipelines. 
This design also amortizes context encoding by reusing precomputed encoder states across queries. 
On question-answering benchmarks, INTRA outperforms strong engineered retrieval pipelines on both evidence recall and end-to-end answer quality. 
Our results demonstrate that attention-based models already
possess a retrieval mechanism that can be elicited, rather than added as an
external module. 

\end{abstract}

\section{Introduction}

\subsection{Motivation}
Large language models are increasingly used in settings where the information
needed to answer a query is sparse relative to the full available corpus. This is the regime in which retrieval-augmented generation (RAG) has become the default design:
a retriever selects candidate evidence, which is then used by a generator to produce an answer \citep{lewis2021retrievalaugmentedgenerationknowledgeintensivenlp}. 
This decomposition is practical because naively concatenating all available context
into a single long prompt is computationally expensive, and even large-context
models remain brittle when the relevant evidence is sparse and distributed
\citep{yen2024helmetevaluatelongcontextlanguage, modarressi2025nolimalongcontextevaluationliteral}.

At the same time, this standard framing encourages a strong architectural
separation between retrieval and generation. The retriever operates over indexed
text or embeddings, while the language model consumes the selected evidence only
after that selection is complete. In practice this
modularity is often helpful, but it can obscure an important fact: attention is
already a query-conditioned mechanism for selecting and weighting relevant
information. This motivates the central question of the paper: can a single
pretrained encoder-decoder retrieve the needed evidence and use it to answer a
query? More broadly, how much of RAG can be handled inside the model itself,
without a separate retriever?

\subsection{Retrieval as an intrinsic capability}
We study question answering using a fixed knowledge base and ask
whether a pretrained encoder-decoder can retrieve, prioritize, and use evidence
drawn from its own representation space. Our central hypothesis is that
pretrained attention-based models already possess an intrinsic retrieval
capability in this setting. We call this regime \emph{INTRA}
(\emph{INTrinsic Retrieval via Attention}): rather than relying on an external retriever, the model selects evidence and generates
answers over the representations produced by its own encoder.

The connection between attention and retrieval is made concrete in
Section~\ref{subsection:attention-retrieval}: both are query-conditioned
matching operations over candidate states. Within this framing, INTRA transforms the decoder’s cross-attention queries into scores for chunk-level retrieval. This perspective does not suggest that attention mechanisms constitute a complete solution for large-corpus retrieval. Rather, it suggests that a pretrained encoder-decoder contains the
right interface for retrieval: query states that express what the decoder needs,
and encoded evidence states that can be selected and consumed without translation into another representation space.

This design has several practical advantages. The same encoded chunk states
are used for both evidence selection and answer generation, reducing the
mismatch between a separately trained retriever and the generator it serves.
Because those states are encoder memories, static evidence can be encoded once
and reused across queries instead of being repeatedly packed into a long decoder
context. Finally, the retrieval mechanism can be adapted with lightweight
decoder-side retrieval queries rather than a separately trained retriever, reducing the need for a dedicated retrieval system. 

\begin{figure}[t]
    \centering
    \includegraphics[width=\linewidth]{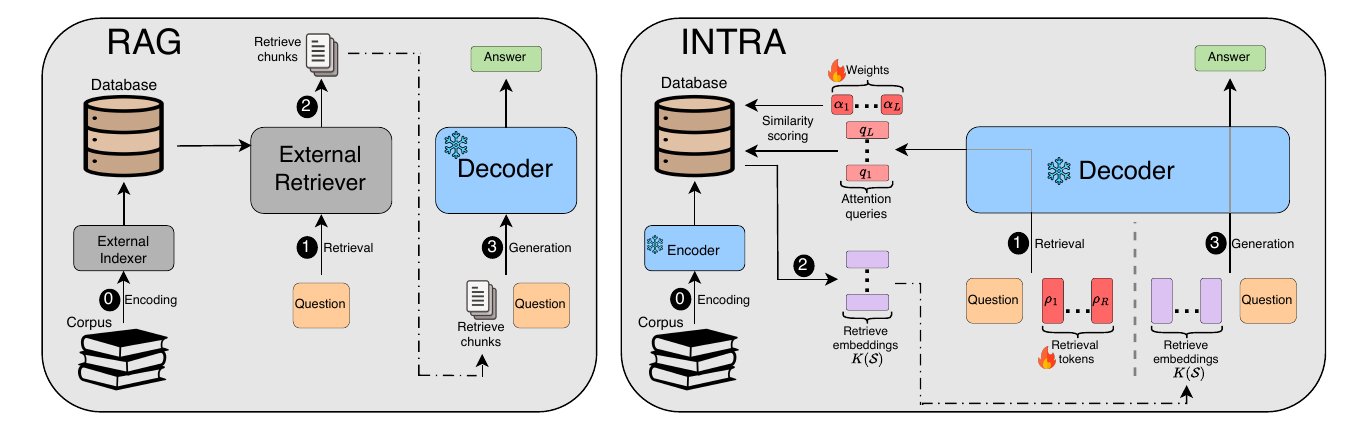}
    \caption{\textbf{Left}: Standard RAG pipeline. An external retriever selects documents which are then re-encoded by the decoder
    to produce an answer. Retrieval and generation operate in separate representation spaces.
    \textbf{Right}: Our method, INTRA, uses a pretrained frozen encoder-decoder for
    both retrieval and generation. The decoder retrieves relevant chunks through its
    cross-attention queries, augmented with learnable retrieval tokens. The retriever and
    generator share a representation space, allowing pre-encoded evidence to be reused
    across queries. No external retriever is required.
    Numbers indicate the sequence of operations.}
    \label{fig:model}
\end{figure}

\vspace{-0.4em}
\subsection{Contributions}
\begin{itemize}
    \setlength{\itemsep}{1pt}   % Gap between items
    \item We formulate \emph{INTRA}, in which a
    single pretrained encoder-decoder model uses one shared representation space
    to couple evidence selection with answer generation.
    \item We identify a minimal architectural recipe for exposing this
    capability: the pretrained encoder's native chunk representations are
    reused directly, and decoder-side retrieval queries refine evidence without introducing a
    separate retriever or compression model (see Figure~\ref{fig:model}).
    \item We show empirically that this unified retrieval-generation design is especially strong on multi-hop question-answering (QA). It rivals strong engineered retrieval pipelines despite their use of large-scale training data, while utilizing the same latent evidence for both selection and generation.
    \item We characterize the computational profile of this design, including
    the reusable-context regime that emerges when static evidence is encoded
    once and reused across queries.
\end{itemize}

\section{Method}
\subsection{Framework formulation}
\label{subsection:framework}
We consider a retrieval-and-generation setting in which the model generates an
output from a small set of relevant evidence chunks. Let
$\mathcal{T}=\{t_i\}_{i=1}^M$ denote the corpus of text chunks, and let
$\mathcal{S}\subseteq\{1,\ldots,M\}$ denote the selected chunk indices. For a
selected set, the retrieved text context is
\[
T(\mathcal{S}) = \Big[\, t_i : i \in \mathcal{S} \,\Big],
\]
where $[\cdot]$ denotes concatenation. Given an input $x$ (e.g., a question), a
decoder $\mathrm{Dec}$ produces the output $y$ conditioned on this context:
\[
y = \mathrm{Dec}\bigl(x,T(\mathcal{S})\bigr).
\]
In a standard RAG pipeline, the selected set $\mathcal{S}$ is obtained from an
\emph{external} retrieval function,
$\mathcal{S}=\mathrm{retrieve}(x,\mathcal{T})$, and $\mathrm{Dec}$ is usually a
separate LLM that generates from the retrieved text.

We focus on \emph{encoder-decoder} models, where the same pretrained model can
encode evidence and decode the answer. The encoder $\mathrm{Enc}$ maps a text
sequence $t$ to token representations
\[
k = \mathrm{Enc}(t) \in \mathbb{R}^{L_c \times d},
\]
where $d$ is the representation dimension. For each corpus chunk, we write
$k_i=\mathrm{Enc}(t_i)$ and denote the pre-encoded chunk set by
$\mathcal{K}=\{k_i\}_{i=1}^M$. For a selected set $\mathcal{S}$, the encoded
context is
\begin{equation*}
    \KS = \Big[\, k_i : i \in \mathcal{S} \,\Big],
\end{equation*}
where the same concatenation notation is now applied to token representations.
Generation in our setting conditions on the encoded context rather than on raw
retrieved text, so
\[
y = \mathrm{Dec}(x,\KS).
\]
To make the decoder queries explicit, we use a view that isolates
the cross-attention computation. Let
\[
\mathrm{Attention}(q,k,v)
\triangleq
\mathrm{softmax}\!\left(\frac{qk^\top}{\sqrt{d}}\right)v .
\]
During the forward pass that computes $\mathrm{Dec}(x,\KS)$, let $h_0=x$ and
let $q_\ell$ denote the query-side state consumed by cross-attention in decoder
layer $\ell$. The simplified
internal recurrence is
\begin{equation}\label{eq:xattn-query}
\begin{array}{rcl}
q_\ell &=& \Psi_\ell(h_{\ell-1}),\\
z_\ell &=& \mathrm{Attention}(q_\ell,\KS,\KS),\\
h_\ell &=& \Phi_\ell(h_{\ell-1},z_\ell),
\qquad \ell=1,\ldots,L.
\end{array}
\end{equation}
Here $\Psi_\ell$ and $\Phi_\ell$ denotes layer-specific transformation, including residual updates, self-attention, feed-forward transformations, and normalization. The decoder output is then produced from the final internal state, where $\mathrm{Out}$ denotes the model's final text-generation head.
\[
y = \mathrm{Dec}(x,\KS) = \mathrm{Out}(h_L).
\]
We also use $\widetilde{\mathrm{Dec}}$ to denote the same decoder forward pass, with the
intermediate query states exposed:
\[
\forall \ell=1,\ldots,L: q_\ell
=
\widetilde{\mathrm{Dec}}_{\ell}(x, K(\mathcal{S})).
\]

\subsection{Attention-based retrieval}
\label{subsection:attention-retrieval}
Cross-attention already scores decoder-side query states against encoder-side
token representations. We use the same matching signal to rank chunks before
generation. Our goal is to convert the token-level comparison in
\eqref{eq:xattn-query} into a single retrieval score for each encoded chunk
$k_i$.
To obtain these scores with a pre-trained \emph{frozen} decoder, we augment the
input $x$ with $R$ trainable retrieval tokens
$\rho\in\mathbb{R}^{R\times d}$. Given token embeddings
$\{x_j\}_{j=1}^{L_q}$, the retrieval input is
\begin{equation}\label{eq:x-rho}
x_{\mathrm{retrieval}} = \bigl[x_1,\dots,x_{L_q}, \rho_1,\dots,\rho_R \bigr].
\end{equation}

To move from token-level attention scores to chunk-level retrieval scores, we
use a scaled ColBERT-style late-interaction score $\mathrm{MaxSim}$
\citep{colbert}\begingroup\renewcommand{\thefootnote}{\fnsymbol{footnote}}\footnotemark[1]\footnotetext[1]{The factor $1/\sqrt{d}$ has no effect on MaxSim ranking; it is included only to make the similarity to the attention score explicit.}\endgroup.
For sequences $u \in \mathbb{R}^{L_u \times d}$ and
$v \in \mathbb{R}^{L_v \times d}$,
\begin{equation*}
    \mathrm{MaxSim}(u, v)
    \triangleq
    \sum_{a=1}^{L_u} \max_{1 \le b \le L_v}
    \left(\frac{u v^\top}{\sqrt{d}}\right)_{a,b}\,,
\end{equation*}
MaxSim uses the same scaled token-level dot product as attention, but aggregates
by taking the best-matching chunk token for each query token rather than
applying a softmax over all tokens.

With learned per-layer aggregation weights $\alpha_\ell$, the score for chunk $i$ is
\begin{equation}
\label{equation:intra_maxsim}
    s_i
    \triangleq
    \mathrm{MaxSim}(\sum_\ell{\alpha_{\ell} q_\ell}, k_i)\quad \mathrm{where}   \quad q_\ell=\widetilde{\mathrm{Dec}}_{\ell}(x_{\mathrm{retrieval}}, K(\mathcal{S}_0)) \, ,
\end{equation}
where $\mathcal{S}_0$ is the initial chunk selection providing the decoder with useful starting context. $\mathcal{S}_0$ is constructed with training-free BM25 ranking \cite{robertson2009bm25}, see Appendix~\ref{sec:context-init} for other choices. 

We then select the chunks with the largest scores:
\begin{equation}\label{eq:s_intra_topk}
    \mathcal{S}_{\mathrm{INTRA}}
    =
    \left\{ i \in \{1,\ldots,M\} : s_i \text{ is among the top-}n \text{ scores} \right\},
\end{equation}
and the ordinary decoder generates conditioned on $\mathcal{S}_{\mathrm{INTRA}}$:
\begin{equation}
    y = \mathrm{Dec}\left(x, K(\mathcal{S}_{\mathrm{INTRA}})\right).
\end{equation}

Inference thus consists of two stages: a retrieval pass $\widetilde{\mathrm{Dec}}(x_{\mathrm{retrieval}}, K(\mathcal{S}_0))$ that exposes the query states $\{q_\ell\}$ used in \eqref{equation:intra_maxsim} to score all chunks, followed by an autoregressive generation utilizing the selected context $K(\mathcal{S}_{\mathrm{INTRA}})$. Because scoring is over the full corpus, $\mathcal{S}_{\mathrm{INTRA}}$ can include chunks outside $\mathcal{S}_0$, unlike reranking methods which only reorder a fixed candidate set.

\section{Practical Implementation}
We now describe the practical changes needed to adapt pretrained encoder-decoder attention models to the INTRA framework. Our implementation starts from T5Gemma2 \citep{zhang2025t5gemma} and modifies the decoder cross-attention so that pre-encoded chunk states can be reused directly for retrieval and generation.

\subsection{Shared context representations across layers}
In T5Gemma2, as in other Transformer-based encoder-decoder models, the cross-attention computation $z_\ell = \mathrm{Attention}(q_\ell, \KS, \KS)$ defined in Eq.~\ref{eq:xattn-query} does not take dot products directly against the raw stored encoder states $\KS$. Instead, the key inputs to the attention function are subject to layer-specific transformations. The corresponding keys are typically computed by applying an RMSNorm with learned scale $\gamma_{K,\ell}$ and a linear projection matrix $W_{K,\ell}$. Thus, in Eq.~\ref{eq:xattn-query} and related equations, we would need to replace $\KS$ with $K_{\ell}(\mathcal{S}) = (\mathrm{RMSNorm}(\KS) \odot \gamma_{K,\ell}) W_{K,\ell}$. This would require computing layer-specific representations $K_{\ell}(S)$ to evaluate MaxSim for retrieval, rather than evaluating against a single reusable context across all layers.

To avoid this overhead, we propose reversed query-key projection \emph{Reverse-QWK} (or \emph{RQWK}), a novel technique that stores one normalized encoder representation $\bar{K}(\mathcal{S})=\mathrm{RMSNorm}(\KS)$ and moves the learned key scale $\gamma_{K,\ell}$ and projection matrix $W_{K,\ell}$ to the query side, defining a modified query transformation:
\begin{equation*}
    \widetilde{q}_{\ell}
    = (q_{\ell} W_{K,\ell}^\top) \odot \gamma_{K,\ell} .
\end{equation*}
Cross-attention can then be computed against the same normalized encoder states for all layers, maintaining equivalence with Eq.~\ref{eq:xattn-query}:

\begin{equation}
\label{equation:attn_reverse_qwk}
    z_\ell = \mathrm{Attention}_{\mathrm{RQWK}}(\widetilde{q}_{\ell},\bar{K}(\mathcal{S}), \bar{K}(\mathcal{S}))
    =
    \mathrm{softmax}\left(\frac{\widetilde{q}_{\ell}\bar{K}(\mathcal{S})^\top}{\sqrt{d}}\right) \bar{K}(\mathcal{S})
\end{equation}
The MaxSim score in Eq.~\ref{equation:intra_maxsim} is computed using these same quantities, $\mathrm{MaxSim}(\widetilde{q}_{\ell}, \bar{k}_i)$ (where $\bar{k}_i = \mathrm{RMSNorm}(k_i)$), so retrieval and attention operate in the same representation space. This preserves the attention scores while allowing retrieval queries from different decoder layers to operate on the same stored chunk pool $\KS$. 

We use Reverse-QWK only as an implementation device; the full derivation including per-head treatment under Grouped-Query Attention, handling of positional embeddings, and the resulting memory savings are deferred to Appendix~\ref{appendix:reverse_qwk}.

\subsection{Retrieval training objective}

Let $\mathcal{O}(x) \subseteq \{1,\dots,M\}$ denote the oracle chunks
for input $x$ (e.g., a question). When retrieval supervision is available, we train the
retrieval scores $s_i$ from \eqref{equation:intra_maxsim} with a cross-entropy objective that places uniform target mass on $\mathcal{O}(x)$, with scores normalized over a candidate subset $\mathcal{C}(x)$:
\[
\mathcal{L}_{\mathrm{retrieval}}
= - \frac{1}{|\mathcal{O}(x)|} \sum_{j \in \mathcal{O}(x)}
\log \frac{\exp(s_j/\tau)}{\sum_{n \in \mathcal{C}(x)} \exp(s_n/\tau)},
\]
where $\tau$ is a temperature parameter. The candidate subset is $\mathcal{C}(x) = \mathcal{O}(x) \cup \mathcal{N}(x)$, where $\mathcal{N}(x)$ holds hard negatives for $x$ and additional in-batch chunks (see Appendix~\ref{sec:negatives} for details). When scoring the full corpus is feasible, we simply take $\mathcal{C}(x)=\{1,\dots,M\}$; subsampling is needed only when $M$ makes this prohibitive. With the decoder frozen, this objective updates
the retrieval tokens $\rho$ in \eqref{eq:x-rho} and the layer aggregation weights $\alpha$ in \eqref{equation:intra_maxsim},
teaching the induced decoder queries to place probability mass on the oracle
evidence set.

\subsection{Approximate similarities with pooled chunk embeddings}\label{sec:chunk-encoding}

Computing $\mathrm{MaxSim}$ against every token is expensive when chunk length $L_c$ is large. For efficient scoring, we replace each encoded chunk $k_i \in \mathbb{R}^{L_c \times d}$ with a fixed-length mean-pooled sequence $\widehat{k}_i \in \mathbb{R}^{L_p \times d}$ where $L_p \ll L_c$. Retrieval scores are computed using $\widehat{k}_i$ in place of $k_i$.
This approximation is natural for INTRA because the pooled vectors are fixed averages of the model's own encoder states, requiring no additional compressor or compression-specific training (distinct from latent-compression approaches \citep{he2026clarabridgingretrievalgeneration}).

We treat $L_p$ as a quality-versus-cost tradeoff. The number of scored vectors, and with it both the scoring cost of $\mathrm{MaxSim}$ and the size of the retrieval index, grows linearly in $L_p$, while retrieval quality improves monotonically with $L_p$ up to the no-pooling limit $L_p \approx L_c$, at which $\widehat{k}_i = k_i$ and the score in \eqref{equation:intra_maxsim} is computed exactly. See Appendix~\ref{appendix:pooled_length} for an ablation of $L_p$.

\section{Benchmarks and Experimental Setup}\label{sec:benchmarks}

We evaluate INTRA on four Wikipedia-based QA benchmarks: HotPotQA
\citep{yang2018hotpotqa}, 2WikiMultihopQA \citep{ho2020constructing},
MuSiQue \citep{trivedi2022musique}, and Natural Questions
\citep{kwiatkowski2019natural}. Together they span bridge and comparison
reasoning, cleaner two-hop evidence chains, compositionally harder multi-hop
questions, and single-hop open-domain QA. We build one shared retrieval candidate pool for all four benchmarks under a
fixed budget of approximately 100M tokens, containing $M=759K$ chunks in total. Full details on the context pool construction and split statistics are provided in Appendix~\ref{appendix:clara_dataset_details}.

We compare INTRA against nine retrieval baselines, including sparse lexical methods (TF-IDF \citep{salton1988termweighting}, BM25 \citep{robertson2009bm25}), dense single-vector models (BGE-large \citep{xiao2023cpack}, Qwen3-Embedding-0.6B/4B \citep{zhang2025qwen3embedding}), reranking (Jina reranker \citep{jinaai2024jinarerankerv2}), hybrid RAG (RRF \citep{cormack2009reciprocal}), and a ColBERT-style MaxSim late-interaction baseline \citep{colbert} (details in Appendix~\ref{appendix:baseline_details}). For retrieval, we report complete-evidence recall at $k \in \{5, 10, 20\}$, defined as the fraction of examples where \emph{all} oracle chunks are retrieved. For end-to-end QA, we take the top-5 retrieved chunks, pack their pre-encoded T5Gemma2 representations as cross-attention context, and generate answers with the T5Gemma2 model, reporting exact match (EM) and token-level F1. All experiments use the open retrieval setting.

\paragraph{Decoder Warm-Start.} The available T5Gemma2 checkpoint is not instruction-tuned, so it can't answer questions out of the box. We first warm-start its decoder on the CLaRa pretraining data \citep{he2026clarabridgingretrievalgeneration}, with the encoder frozen. We use the Reverse-QWK parameterization of \eqref{equation:attn_reverse_qwk}, in which the decoder attends to a shared pool of representations $\{\bar{k}_i\}_{i=1}^M$. Training adapts the decoder to this pool, which is then stored and reused at inference. The objective is next-token cross-entropy on answer tokens.

\paragraph{Retrieval training.} During retrieval training, the encoder-decoder backbone is frozen, and only the retrieval tokens $\rho_i$ and layer aggregation weights $\alpha_l$ are optimized, for a total of approximately $164$K parameters. Initial context $\mathcal{S}_0$ uses $n_0=10$ and pooled chunks of length $L_p=7$.

\paragraph{Evaluation.} At evaluation, QA builds a five-chunk context: the top four retrieved chunks from $\mathcal{S}_{\mathrm{INTRA}}$ plus the top initial context chunk from $\mathcal{S}_{0}$. We then generate with deterministic greedy decoding. 

Further details and ablations are reported in 
Appendices~\ref{appendix:implementation_details} and~\ref{appendix:ablation_study}.

\section{Results}
\label{sec:results}

We organize the results around the four empirical questions that motivate the
paper. First, does INTRA improve retrieval of complete evidence
sets (Section~\ref{sec:retrieval_results})? Second, do those gains translate into better end-to-end answer quality (Section~\ref{sec:generation_results})?
Third, what efficiency advantage appears once chunk representations are
reused rather than re-encoded from raw text (Section~\ref{sec:efficiency_results})? Finally, do the retrieval gains persist when the corpus grows to a full Wikipedia scale (Section~\ref{sec:full_wiki_results})?

\subsection{Retrieval Results}
\label{sec:retrieval_results}

\begin{figure}[t]
    \centering
    \includegraphics[width=\linewidth]{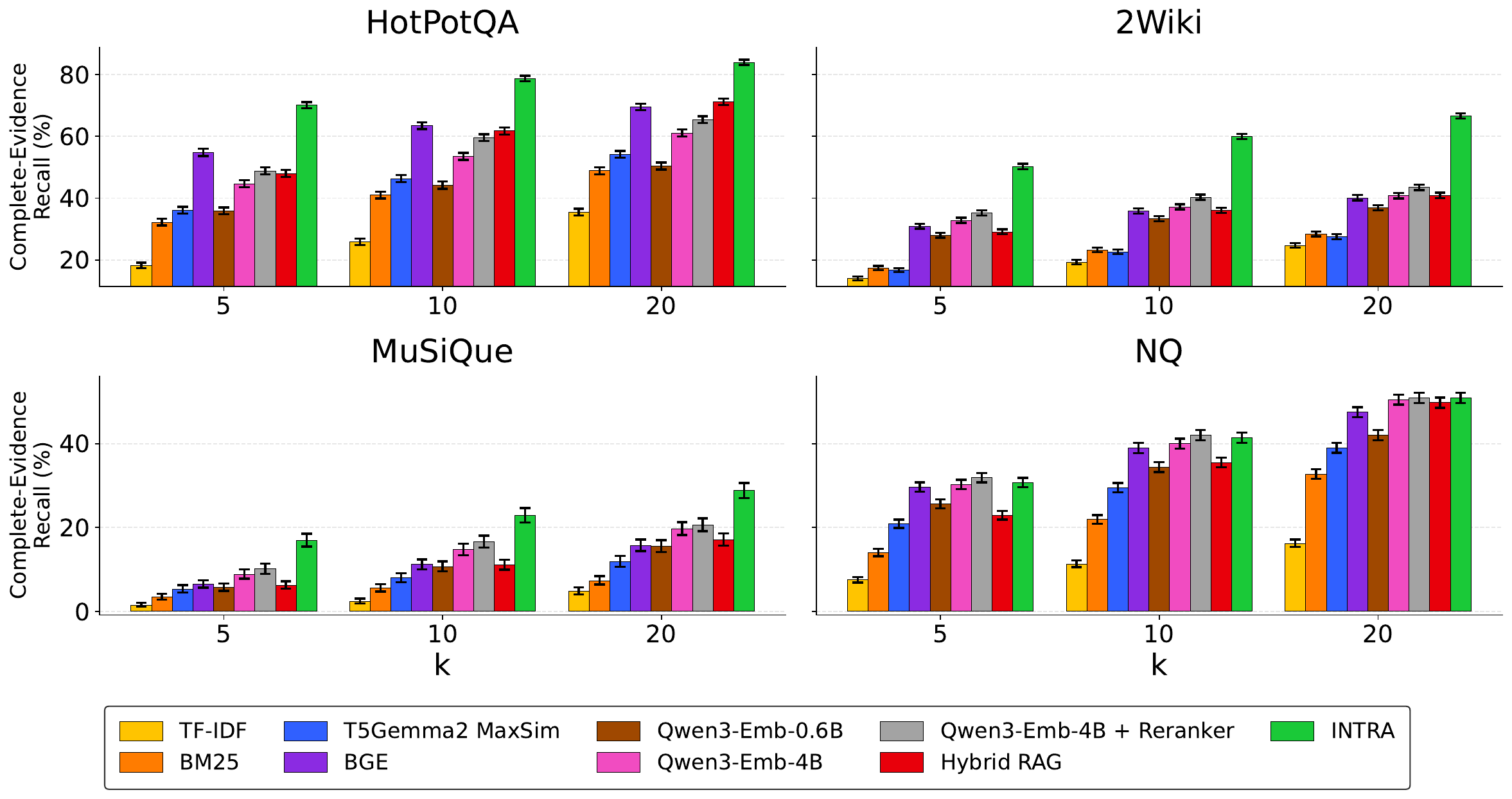}
    \caption{Complete-evidence recall: the percentage of examples for which \emph{all} supporting facts are retrieved. 
    INTRA performs best on multi-hop benchmarks (HotPotQA, 2Wiki, MuSiQue) that require evidence assembly. 
    NQ’s single-hop nature minimizes this benefit.
    }
    \label{fig:retrieval_recall_all_oracles}
\end{figure}

Figures~\ref{fig:retrieval_recall_all_oracles} and~\ref{fig:retrieval_recall_average} report the complete-evidence recall and average recall, respectively, for $k \in \{5, 10, 20\}$ across the four evaluation benchmarks.
Complete-evidence recall@k is the fraction of examples for which \emph{all} annotated supporting chunks are retrieved within the top-$k$ results. We view this metric as the clearest proxy for retrieval quality, because it rewards recovering the full supporting set rather than only incomplete supporting evidence.

The main pattern is that INTRA is strongest on multi-hop benchmarks that
require assembling multiple pieces of evidence (HotPotQA, 2Wiki, MuSiQue).
INTRA ranks chunks by the decoder's own attention queries, which reflect what
the decoder needs to generate the answer. A leave-one-out analysis on HotPotQA
(Appendix~\ref{appendix:attribution}) supports this: the higher INTRA ranks a
chunk, the more removing it from the context degrades the generator's answer.
This matters most for multi-hop questions, where a \emph{bridge} passage links
the question to an \emph{answer-bearing} passage that often shares little
wording with the question. A question-similarity retriever tends to find the
bridge, whereas INTRA prioritizes the answer-bearing passage and recovers both more often ($64.1\%$ vs.\
$54.3\%$ at $k=10$, Appendix~\ref{appendix:attribution}). On NQ, retrieval often reduces to finding one directly
supporting passage, leaving less room for this advantage. See Appendix~\ref{appendix:additional_results} for further results, analyses, and ablations.

\begin{figure}[t]
    \centering
    \includegraphics[width=\linewidth]{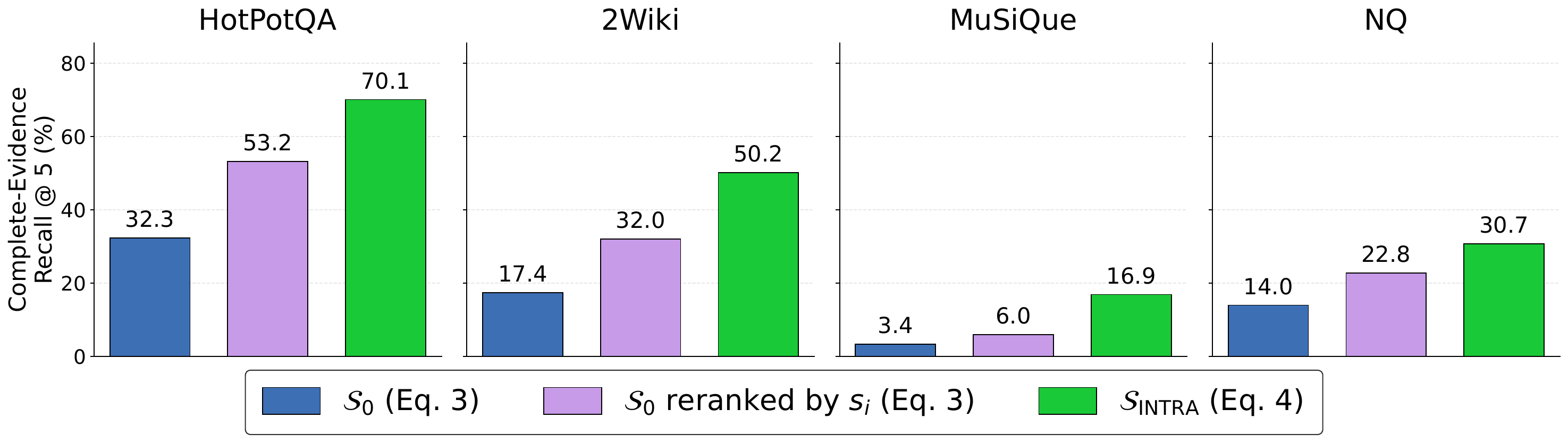}
    \caption{Complete-evidence recall@5 for the initial set $\mathcal{S}_0$, $\mathcal{S}_0$ reranked, and the final set $\mathcal{S}_{\mathrm{INTRA}}$. INTRA performs corpus-wide scoring and can recover evidence outside the initial candidate pool.
    }
    \label{fig:retrieval_recall_initial_ranked_final}
\end{figure}

In Figure~\ref{fig:retrieval_recall_initial_ranked_final} we also compare three top-5 evidence sets: the initial retrieval set $\mathcal{S}_0$, the same initial set
reranked by the decoder scores $s_i$ from
Eq.~\ref{equation:intra_maxsim}, and the final INTRA set
$\mathcal{S}_{\mathrm{INTRA}}$ from Eq.~\ref{eq:s_intra_topk}.
The results show that reranking $\mathcal{S}_0$ is beneficial, but full-corpus INTRA scoring yields the largest gains by recovering evidence absent from the initial pool.

\subsection{End-to-End Question-Answering Results}
\label{sec:generation_results}

Table~\ref{tab:e2e_metrics_at_5_t5gemma} evaluates the end-to-end retrieval-and-generation behavior of INTRA. 
We vary the retrieval method while keeping the same T5Gemma2 generator, reporting both exact match (EM) and token-level F1 (full results are in Appendix~\ref{appendix:additional_results}).
INTRA surpasses all baselines on multi-hop benchmarks (HotPotQA, 2Wiki, MuSiQue), consistent with the retrieval results of Section~\ref{sec:retrieval_results}. 
This is notable because INTRA's retrieval signal comes from a frozen decoder pretrained only for generation, whereas baselines such as BGE and Qwen-Embedding are pretrained for retrieval on large-scale retrieval corpora (that include HotPotQA and NQ as supervision \cite{thakur_bge_full_data, zhang2025qwen3embedding}).

\begin{table*}[t]
    \centering
    \small
    \caption{End-to-end question-answering performance across retrieval methods with a fixed T5Gemma2 generator. 
    INTRA performs best on all multi-hop benchmarks.
    }
    \label{tab:e2e_metrics_at_5_t5gemma}
    \begin{tabular}{lcccccccccc}
        \toprule
        & \multicolumn{2}{c}{HotPotQA} & \multicolumn{2}{c}{2Wiki} & \multicolumn{2}{c}{MuSiQue} & \multicolumn{2}{c}{NQ} & \multicolumn{2}{c}{Average} \\
        \cmidrule(lr){2-3} \cmidrule(lr){4-5} \cmidrule(lr){6-7} \cmidrule(lr){8-9} \cmidrule(lr){10-11}
        Retrieval method & EM & F1 & EM & F1 & EM & F1 & EM & F1 & EM & F1 \\
        \midrule
        TF-IDF & 34.2 & 44.5 & 39.0 & 42.7 & 5.3 & 14.6 & 34.9 & 42.9 & 28.4 & 36.2 \\
        BM25 & 40.5 & 52.0 & 41.7 & 45.2 & 7.7 & 16.8 & 43.4 & 51.8 & 33.3 & 41.5 \\
        MaxSim & 40.7 & 52.2 & 41.6 & 45.8 & 10.1 & 19.8 & 48.4 & 57.8 & 35.2 & 43.9 \\
        Hybrid RAG & 43.4 & 54.3 & 46.0 & 49.9 & 10.6 & 20.1 & 50.5 & 59.1 & 37.6 & 45.8 \\
        BGE & 41.9 & 53.0 & 46.1 & 49.6 & 10.8 & 19.7 & 52.2 & 61.2 & 37.8 & 45.9 \\
        Qwen3-Emb-0.6B  & 37.0 & 47.7 & 45.7 & 49.8 & 11.1 & 20.0 & 36.4 & 44.6 & 32.6 & 40.5 \\
        Qwen3-Emb-4B & 40.3 & 51.2 & 46.0 & 50.1 & 12.7 & 21.8 & 54.5 & 63.7 & 38.4 & 46.7 \\
        Qwen3-Emb-4B + Reranker & 41.6 & 53.6 & 46.8 & 50.8 & 13.3 & 22.5 & \textbf{55.1} & \textbf{64.2} & 39.2 & 47.5 \\
        INTRA & \textbf{48.8} & \textbf{60.6} & \textbf{52.3} & \textbf{56.4} & \textbf{13.9} & \textbf{23.3} & 53.7 & 63.4 & \textbf{42.2} & \textbf{50.9} \\  
        \bottomrule
    \end{tabular}
\end{table*}

Table~\ref{tab:gap_closure} compares using a shared decoder for retrieval and generation against coupling an INTRA retriever with a stronger generator. While superior reasoning and parametric knowledge allow more capable generators to boost performance, INTRA retrieval enhances performance by aligning evidence with the decoder’s specific attention patterns. To isolate the impact of generator strength, we measure how much of the EM gap between random and complete-evidence contexts is closed by INTRA:
\begin{equation*}
    \mathrm{GapClosure}
    =
    100 \cdot
    \frac{
        \mathrm{EM}(\mathrm{INTRA}) -
        \mathrm{EM}(\mathrm{random})
    }{
        \mathrm{EM}(\mathrm{complete}) -
        \mathrm{EM}(\mathrm{random})
    },
\end{equation*}
where the parenthetical term
indicates whether generation uses chunks from $\mathcal{S}_{\mathrm{INTRA}}$, random chunks, or
the complete-evidence (oracle) chunks.
Utilizing the same T5Gemma2 decoder for both retrieval and generation closes the largest average gap, demonstrating the benefit of coupling the retriever and generator. This highlights the need for stronger INTRA backbones, given that open-source encoder-decoder models are currently scarcer and weaker than decoder-only options.

\begin{table*}[t]
    \centering
    \small
    \caption{Generator compatibility with a T5Gemma2-INTRA retriever.
    Gap closure measures the percentage of the EM gap from random chunks to complete-evidence chunks recovered by INTRA retrieval.
    Sharing a decoder across retrieval and generation aligns the retrieved evidence with the generator's attention, closing the largest gap.}
    \label{tab:gap_closure}
    \begin{tabular}{lccccc}
        \toprule
        Generator & HotpotQA & 2Wiki & MuSiQue & NQ & Average \\
        \midrule
        Mistral0.3-7B & 69.7\% & 52.4\% & 27.5\% & 73.6\% & 55.8\% \\
        Llama3.1-8B & 70.1\% & 51.8\% & 29.5\% & 71.9\% & 55.8\% \\
        Phi-4-mini & 68.3\% & 37.9\% & 23.9\% & 74.6\% & 51.2\% \\
        Gemma-4-E2B & 68.3\% & 55.7\% & 27.7\% & 75.3\% & 56.8\% \\
        Qwen3.5-9B & 71.7\% & 57.0\% & 30.0\% & \textbf{76.3\%} & 58.8\% \\
        Qwen3.5-27B & 71.8\% & 58.0\% & 29.6\% & 72.6\% & 58.0\% \\
        T5Gemma2-4B (same as retriever) & \textbf{73.1\%} & \textbf{59.8\%} & \textbf{33.1\%} & 74.7\% & \textbf{60.2\%} \\
        \bottomrule
    \end{tabular}
\end{table*}

\subsection{Efficiency Results}\label{sec:efficiency_results}
INTRA's encoder-decoder design also yields a direct efficiency benefit. Standard
RAG typically retrieves text, so after retrieval the generator re-encodes the
selected chunks before decoding. INTRA retrieves pre-encoded chunks from
$\mathcal{K}$ instead, and those states feed into generation as decoder
cross-attention memory. Retrieval and generation incur their usual costs\footnote{Retrieval complexity scales as $\bigO(\sqrt{M} L_q L_c)$ in practice using inverted file (IVF) approximate nearest-neighbor (ANN) search e.g.~with FAISS or cuVS, \citep{colbert,Johnson2019,cuVS2026}.},
but the selected evidence is no longer re-encoded at query time. Table~\ref{tab:complexity_professional} summarizes this computational trade-off (detailed analysis in Appendix~\ref{appendix:compute_analysis}).

\begin{table}[t]
\centering
\caption{Computational trade-offs across full-context prompting,
standard RAG, and INTRA. Variables denote number of corpus chunks $M$, chunk length $L_c$, query length $L_q$, retrieved chunks $k$, and generation length $L_g$. INTRA has the same retrieval and generation terms as
RAG, but avoids re-encoding retrieved evidence during prefilling when chunk
representations are reusable across queries. In our setting where $M \gg k$, full-context prefilling is computationally infeasible.}
\
\label{tab:complexity_professional}
\small{
\begin{NiceTabular}{@{}lcccc@{}}
    \CodeBefore
        % This creates a solid block from row 3, col 4 to row 4, col 4
        \rectanglecolor{gray!25}{3-4}{4-4} 
    \Body
    \toprule
    \textbf{Model} & \textbf{Pre-Query} & \textbf{Retrieval} & \textbf{Prefilling} & \textbf{Generation} \\
    \midrule
    % \textbf{Full Context} & $\bigO(1)$ & $\bigO(1)$ & $\bigO((L_q + N)^2)$ & $\bigO(L_g(L_q + N + L_g))$ \\
    \textbf{Full Context} & $\bigO(1)$ & $\bigO(1)$ & $\bigO((L_q + ML_c)^2)$ & $\bigO(L_g(L_q + ML_c + L_g))$ \\
    \addlinespace[0.5em]
    \textbf{Standard RAG} & $\bigO(ML_c^2)$ & $\bigO(\sqrt{M} L_q L_c)$ & $\bigO((L_q + kL_c)^2)$ & $\bigO(L_g(L_q + kL_c + L_g))$ \\
    \addlinespace[0.5em]
    \textbf{INTRA} & $\bigO(ML_c^2)$ & $\bigO(\sqrt{M} L_q L_c)$ & $\bigO(L_q(L_q+kL_c))$ & $\bigO(L_g(L_q + kL_c + L_g))$ \\
    \bottomrule
\end{NiceTabular}
}
\end{table}

\begin{figure}[t]
    \centering
    \includegraphics[width=\linewidth]{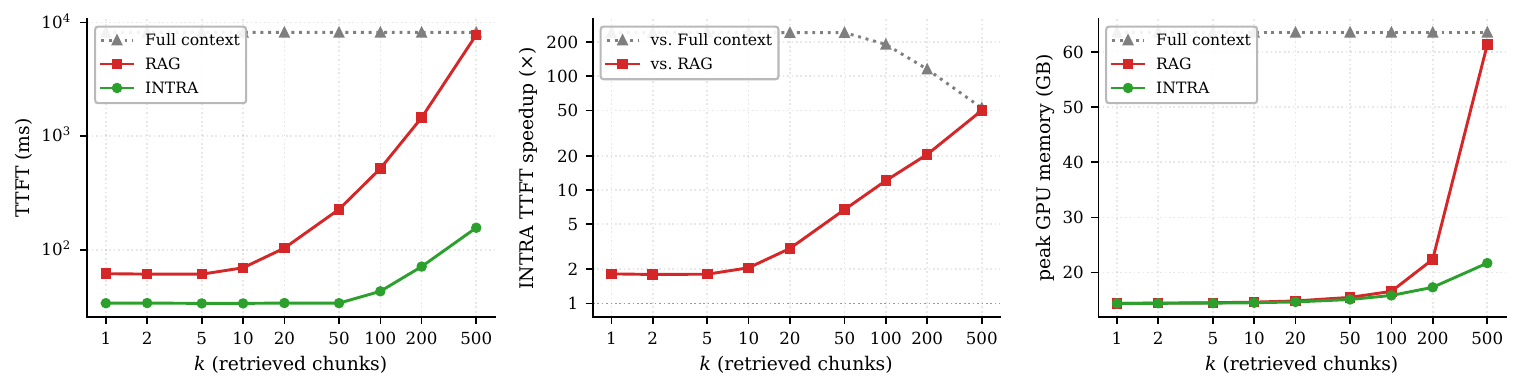}
    \caption{Time to first token vs. number of retrieved chunks $k$, excluding retrieval time. INTRA reuses pre-encoded evidence, while standard RAG re-encodes the retrieved text before decoding.
    }
    \label{fig:bench_ttft_k}
\end{figure}

We isolate this effect with a time-to-first-token (TTFT) benchmark in
Figure~\ref{fig:bench_ttft_k}, which measures the generator-side cost after evidence has
been selected.  As the number of retrieved chunks $k$ grows, INTRA's prefill time grows linearly because it cross-attends to stored encoder states, whereas standard RAG's grows quadratically because it re-encodes the chunks.
Appendix~\ref{appendix:benchmark_details} provides the setup and further generation throughput measurements.

INTRA avoids re-encoding the retrieved chunks at a modest storage cost. Because $\bar{K}(\mathcal{S})$ has no layer or head axis, it stores one $d$-dimensional vector per token, about 2.5~TB for a 1-billion-token corpus at 8-bit precision (Appendix~\ref{appendix:memory_efficiency}).

\subsection{Retrieval at Full-Wikipedia Scale}\label{sec:full_wiki_results}

\begin{figure}[t]
    \centering
    \includegraphics[width=0.95\linewidth]{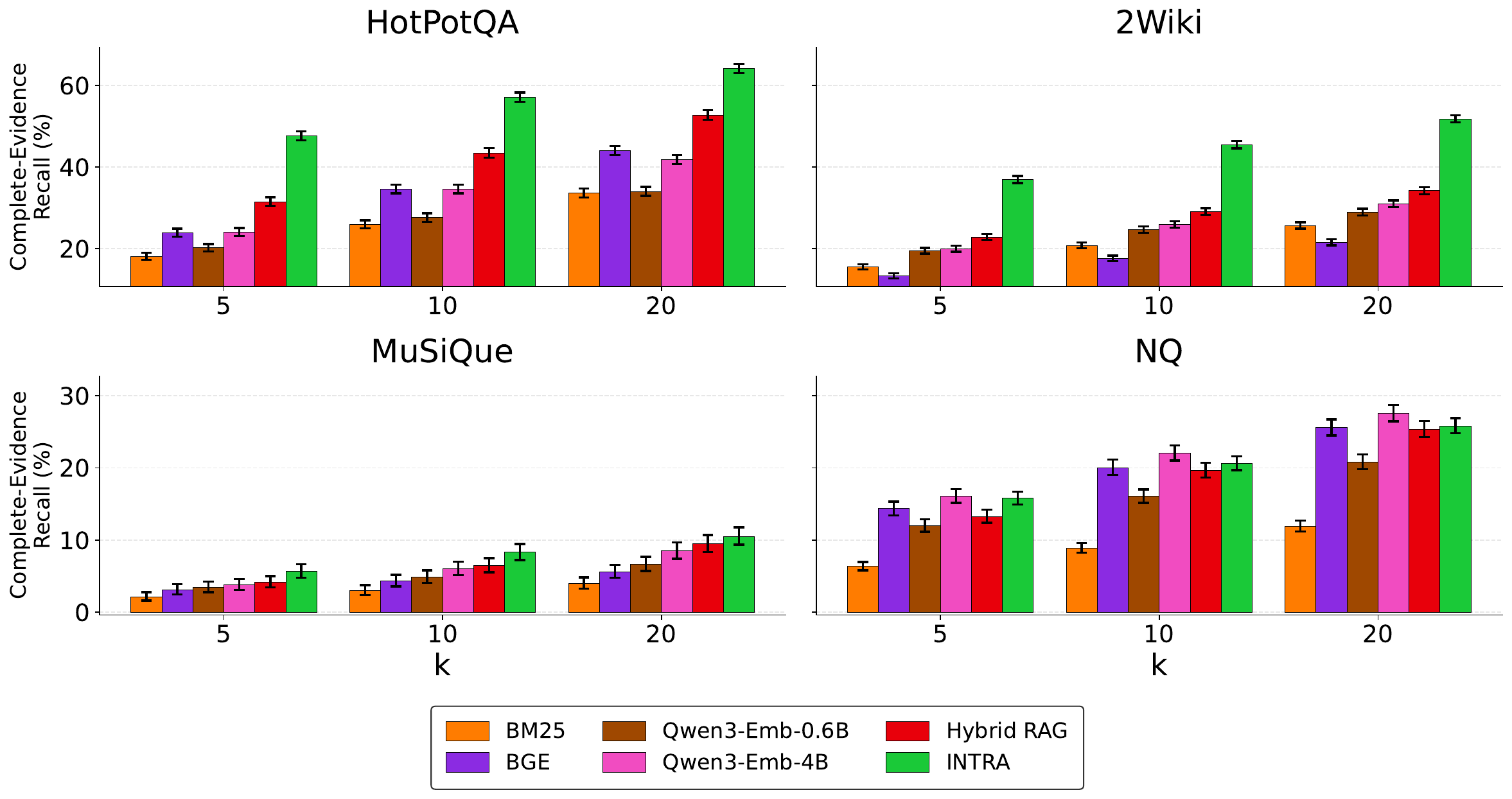}
    \caption{
    Complete-evidence recall on the full Wikipedia corpus
    (21M chunks, 3B tokens), 30$\times$ larger than the pool in
    Figure~\ref{fig:retrieval_recall_all_oracles}. INTRA's advantage on multi-hop
    benchmarks persists at this scale.
    }
\label{fig:retrieval_recall_all_oracles_full_wiki}
\end{figure}

Our main experiments use a 100M-token pool, which keeps analyses and ablations tractable. A natural concern is whether INTRA remains discriminative as the number
of distractors grows. To address this, we verify INTRA on the full Wikipedia corpus \cite{karpukhin-etal-2020-dense}, which consists of around 21M chunks (a total of ${\sim}$3B tokens). Figure~\ref{fig:retrieval_recall_all_oracles_full_wiki} shows the Complete-evidence recall. As before, INTRA outperforms all baselines on multi-hop tasks, showing that its advantage persists as the corpus grows.

\section{Related Work}

\paragraph{Retrieval-Augmented Generation Pipelines.}
Most knowledge-intensive QA systems adopt a modular retrieval-then-generation architecture.
REALM incorporates retrieval into language-model pretraining \citep{realm}, and RA-DIT later
instruction-tunes both retriever and generator \citep{lin2024radit}. DPR established dense passage retrieval as a strong open-domain QA primitive \citep{karpukhin-etal-2020-dense}, and
RAG couples a retriever with a sequence-to-sequence generator
\citep{lewis2021retrievalaugmentedgenerationknowledgeintensivenlp}.  
Atlas \citep{atlas} jointly pre-trained a retriever with an encoder-decoder generator, to optimize few-shot learning from unstructured text. 
More recently, CLaRa \citep{he2026clarabridgingretrievalgeneration} compresses documents into retrievable latent vectors, and jointly optimizes reranking and generation.
%yet it still relies on an external retrieval model.
INTRA differs from CLaRa along three axes: (i) it reuses the model's native representation space, (ii) it performs full-corpus scoring, and (iii) it is built upon a frozen encoder-decoder model.

Multi-pass agentic RAG systems instead interleave reasoning and repeated retrieval \citep{li2025searcho1agenticsearchenhancedlarge,asai2023selfrag,singh2026agenticretrievalaugmentedgenerationsurvey}. INTRA targets the single-pass retrieval block that could be used within such pipelines, rather than the pipeline-level agentic loop itself. A related line of work retrieves iteratively while drawing on side information such as Wikipedia's hyperlink graph and relies on extractive span predictors for answer extraction \citep{zhu2021aiso,ma2023chainofskills,zhang2021tprr,li2021hopretriever}, whereas we target the case of single-pass retrieval over a plain passage corpus paired with a generative reader; absolute scores are therefore not directly comparable.

\paragraph{Late Interaction and Representation-Space Retrieval.}
Our retrieval formulation is close to late-interaction systems such
as ColBERT \citep{colbert}, ColBERTv2 \citep{colbertv2} and ColPali \citep{faysse2024colpaliefficientdocumentretrieval}, which compare query and document tokens via MaxSim-style matching over multi-vector representations. Whereas late-interaction systems rely on a dedicated retriever to score query-document matches, INTRA lets the decoder's own cross-attention perform this matching and then consume the matched representations during generation.

\paragraph{Memory, Latent Retrieval, and Unified Retrieval-Generation.}
Attention has long been viewed as content-based memory access. 
Memory Networks \citep{memory_networks} framed QA as differentiable lookup, while RETRO and InstructRetro \citep{retro, wang2024instructretroinstructiontuningpost} inject retrieved chunks into the model's computation. 
Similarly, Titans \citep{behrouz2026titans} argues that long-context modeling requires explicit memory mechanisms rather than larger attention windows. 
This intuition is compatible with our perspective, yet they differ in the source of memory: Titans learns one, while INTRA reuses the model's activations over an evidence pool.

\paragraph{Long-Context Modeling and Efficient Sequence Architectures.}
Transformer architecture dominance for language modeling can be largely attributed to self-attention, which provides flexible token interactions \citep{NIPS2017_3f5ee243}.
However, dense attention's quadratic cost hinders long-context scaling, prompting sparse and linear alternatives \citep{beltagy2020longformerlongdocumenttransformer,
zaheer2021bigbirdtransformerslonger,
choromanski2022rethinkingattentionperformers,
katharopoulos2020transformersrnnsfastautoregressive}.
More recent models, such as Mamba, explore further reductions in context-processing cost
\citep{mamba,mamba2,poli2023hyenahierarchylargerconvolutional,
lieber2024jambahybridtransformermambalanguage}.
However, recent long-context
benchmarks suggest that merely increasing the window size does not ensure
robust evidence use when relevant information is sparse
\citep{yen2024helmetevaluatelongcontextlanguage,
modarressi2025nolimalongcontextevaluationliteral}. Our goal is therefore not to
replace long-context modeling, but to study a setting in which sparse evidence
must be identified and used reliably.

\section{Limitations}

Our experiments focus on a fixed context pool. 
We discuss the practicality of a billion-token corpus in Appendix~\ref{appendix:memory_efficiency}, 
but do not position INTRA as a replacement for RAG in open-web retrieval or web-scale settings. 
Likewise, while we show that a pretrained encoder-decoder exhibits strong retrieval behavior in this regime, 
we do not show that this mechanism generalizes to a general-purpose retriever.
Moreover, we focus on a single implementation family: a T5Gemma2-style encoder-decoder with Reverse-QWK, 
evaluated on text-QA benchmarks with short answers. INTRA's reliance on encoder-decoder cross-attention 
also excludes decoder-only models. Extending these findings across scales, modalities, dynamic corpora, 
and architectures is left for future work.

The systems evaluation isolates a single mechanism. TTFT in Section~\ref{sec:efficiency_results} excludes retrieval time to isolate the cost of re-encoding versus reuse. Deployment cost also depends on indexing, 
storage format, and data movement, and token-level memories can be substantially larger than compressed 
retrieval indices, so end-to-end trade-offs may differ from those analyzed here.

\section{Conclusion}

Our central conclusion is conceptual: retrieval capabilities, typically outsourced to a separate module, can be carried out within an encoder-decoder's representation space. We introduce INTRA, a framework that unifies retrieval and generation by eliciting the intrinsic retrieval mechanism of attention-based models. By utilizing the same pretrained model for both tasks, INTRA eliminates the representation mismatch between the retriever and generator.

Empirically, INTRA shows strong performance on multi-hop question-answering benchmarks (HotPotQA, 2WikiMultihopQA, and MuSiQue),
 achieving results competitive with or exceeding several engineered RAG pipelines in both complete-evidence recall and end-to-end answer quality. 
 Moreover, this shared-space formulation can yield a computational advantage: static evidence can be encoded once and reused across queries, 
 reducing prefilling costs and time-to-first-token during generation. 
 Ultimately, these findings suggest that attention-based encoder-decoders may offer a promising, unified alternative to traditional modular RAG architectures.

\newpage

\begin{ack}

The research of DS was funded by the European Union (ERC, A-B-C-Deep, 101039436). Views and opinions expressed are however those of the author only and do not necessarily reflect those of the European Union or the European Research Council Executive Agency (ERCEA). Neither the European Union nor the granting authority can be held responsible for them. The research of DS was also supported by the Council for Higher Education in Israel under the Moonshot Project.  

\end{ack}

% \newpage

\bibliography{neurips_2026}
\bibliographystyle{neurips_2026}

\newpage
\appendix

\begin{center}
    \LARGE\textbf{Supplementary Material} \\ \vspace{0.2cm}
    \Large\textbf{Retrieval from Within: An Intrinsic Capability of Attention-Based Models}
\end{center}
\vspace{0.5cm}

\section{Reverse-QWK}
\label{appendix:reverse_qwk}

In Transformer encoder-decoder models like T5Gemma2, the standard attention logits for decoder layer $\ell$ are evaluated against keys computed from static encoder representations $\KS$:
\begin{equation}
\frac{q_\ell k_\ell^\top}{\sqrt{d}} \quad \text{where} \quad k_\ell = (\bar{k} \odot \gamma_{K,\ell}) W_{K,\ell}, \quad \bar{k} = \mathrm{RMSNorm}(\KS)
\label{eq:standard_score}
\end{equation}
Because the projection $W_{K,\ell}$ and learned scale $\gamma_{K,\ell}$ are layer-specific, evaluating these logits requires computing a separate $K$ for every layer. This makes it difficult to leverage a single approximate nearest neighbor (ANN) index built on the shared encoder representation.

By moving the key projection and learned scale to the query side, Reverse-QWK defines a transformed query:
\begin{equation}
\widetilde{q}_{\ell} = (q_\ell W_{K,\ell}^\top) \odot \gamma_{K,\ell} 
\label{eq:reverse_q}
\end{equation}
Substituting this into the standard logit computation, and letting $\mathbf{\Gamma}_{K,\ell} = \mathrm{diag}(\gamma_{K,\ell})$, we see the logits remain identical:
\begin{align}
q_\ell K^\top
&= q_\ell \left(\bar{k} \mathbf{\Gamma}_{K,\ell} W_{K,\ell}\right)^\top \nonumber\\
&= q_\ell W_{K,\ell}^\top \mathbf{\Gamma}_{K,\ell} \bar{k}^\top \nonumber\\
&= \underbrace{((q_\ell W_{K,\ell}^\top) \odot \gamma_{K,\ell})}_{\widetilde{q}_\ell} \bar{k}^\top.
\label{eq:equivalence}
\end{align}
This allows attention scores to be written as direct dot products $(\widetilde{q}_\ell \bar{k}^\top)/\sqrt{d}$ against one normalized encoder pool $\bar{k}$, enabling a single ANN index to be shared across layers and heads.

\subsection{Dimensionalities and Group-Query Attention}
\label{appendix:dimensions_gqa}

The simplified equations above suppress head structure for clarity. In T5Gemma2 the decoder uses Group-Query Attention (GQA): each layer has $n_h$ query heads but only $n_{\mathrm{kv}}\le n_h$ key/value heads, with replication factor $n_{\mathrm{rep}} = n_h / n_{\mathrm{kv}}$, so each KV head is shared by $n_{\mathrm{rep}}$ Q-heads. With encoder hidden size $d$ and per-head dimension $d_h$, the relevant tensors are
\begin{itemize}
    \item $\KS \in \mathbb{R}^{N \times d}$ and $\bar{k} = \mathrm{RMSNorm}(\KS) \in \mathbb{R}^{N \times d}$: a single, head-agnostic, shared encoder pool;
    \item $W_{K,\ell} \in \mathbb{R}^{d \times n_{\mathrm{kv}} d_h}$, viewed as $n_{\mathrm{kv}}$ per-KV-head blocks $W_{K,\ell}^{(g)} \in \mathbb{R}^{d \times d_h}$;
    \item $\gamma_{K,\ell} \in \mathbb{R}^{d_h}$, the per-head-dim RMSNorm scale shared across KV heads;
    \item $q_\ell \in \mathbb{R}^{L_q \times n_h \times d_h}$, with per-Q-head slices $q_\ell^{(h)} \in \mathbb{R}^{d_h}$.
\end{itemize}
At the per-head level the Reverse-QWK transformation in Eq.~\ref{eq:reverse_q} becomes
\begin{equation*}
\widetilde{q}_\ell^{(h)} \;=\; \bigl( q_\ell^{(h)} \odot \gamma_{K,\ell} \bigr) \, \bigl(W_{K,\ell}^{(g(h))}\bigr)^{\!\top} \;\in\; \mathbb{R}^{d}, \qquad g(h) = \lfloor h / n_{\mathrm{rep}} \rfloor,
\end{equation*}
so each Q-head $h$ uses the $W_K$ block of the KV-group $g(h)$ it belongs to, and lifts $q_\ell^{(h)}$ from $\mathbb{R}^{d_h}$ into the shared encoder space $\mathbb{R}^{d}$. The dot product $\widetilde{q}_\ell^{(h)} \bar{k}^{\!\top}$ inherits the standard per-head attention scale of Eq.~\ref{eq:standard_score} ($1/\sqrt{d_h}$ in the T5Gemma2 implementation).

Because $\bar{k}$ has no head dimension, GQA combines naturally with Reverse-QWK: in standard cross-attention GQA still requires expanding $K$ to $n_h$ heads on the encoder side, costing $\mathcal{O}(N\, n_h\, d_h)$ memory; under Reverse-QWK only the head-agnostic pool $\bar{k}$ is materialized, costing $\mathcal{O}(N d)$ regardless of $n_h$, and the GQA replication of $W_K^{(g)}$ across the $n_{\mathrm{rep}}$ Q-heads of each group happens entirely on the (small) query side.

\subsection{Implementation Details}

The following PyTorch-style pseudocode illustrates the Reverse-QWK cross-attention computation, including the GQA structure described in Section~\ref{appendix:dimensions_gqa}. Compared to the standard T5Gemma2 cross-attention, the only structural change is that $W_k$ and $\gamma_k$ are moved to the query side, and $K$ is replaced by the shared, head-agnostic, $d$-dimensional pool $\bar{k}$ -- $V$ and the rest of attention are unchanged.

{\footnotesize
\begin{verbatim}
def reverse_qwk_attention(x, kv_x, W_q, W_k, W_v, gamma_k):
    # x:        [batch, q_len,  dim]      decoder hidden states
    # kv_x:     [batch, kv_len, dim]      already-RMSNormed encoder pool: kv_bar
    # W_q:      [n_h  * d_h, dim]         nn.Linear weight, n_h  Q-heads
    # W_k, W_v: [n_kv * d_h, dim]         nn.Linear weight, n_kv KV-heads (GQA, n_kv <= n_h)
    # gamma_k:  [d_h]                     per-head-dim K-norm scale (shared across KV-heads)
    # n_rep = n_h // n_kv                 Q-heads per KV-group

    # Standard per-head Q projection (q_norm and RoPE on q omitted for brevity)
    Q = (x @ W_q.T).view(batch, q_len, n_h, d_h)              # [B, q_len, n_h, d_h]

    # Reverse key projection: gamma_k first (per d_h), then W_k.T (per Q-head, GQA-replicated)
    W_k_per_kv = W_k.view(n_kv, d_h, dim)                     # [n_kv, d_h, dim]
    W_k_per_q  = W_k_per_kv.repeat_interleave(n_rep, dim=0)   # [n_h,  d_h, dim] (GQA replication)
    Q_tilde = einsum('bqhd,hdi->bhqi', Q * gamma_k, W_k_per_q)  # [B, n_h, q_len, dim]

    # kv_x is already the shared pool kv_bar = RMSNorm_d(K(S)); one key per token, head-agnostic
    kv_bar = kv_x.unsqueeze(1)                                # [B, 1, kv_len, dim] (broadcast over heads)

    # Scores: dot product over the full embedding dim d, per-head scaled by 1/sqrt(d_h)
    scores = (Q_tilde @ kv_bar.transpose(-2, -1)) / sqrt(d_h) # [B, n_h, q_len, kv_len]

    # V keeps the standard GQA path: project to n_kv heads, then expand to n_h via repeat_kv
    V = (kv_x @ W_v.T).view(batch, kv_len, n_kv, d_h)
    V = repeat_kv(V, n_rep).transpose(1, 2)                   # [B, n_h, kv_len, d_h]

    output = softmax(scores, dim=-1) @ V                      # [B, n_h, q_len, d_h]
    return output                                             # then reshape and apply W_o
\end{verbatim}
}

\subsection{Practical Considerations}
\label{appendix:memory_efficiency}

\textbf{Sharing $\bar{k}$ across layers and heads:} Standard cross-attention requires materializing layer- and head-specific keys $k_\ell^{(g)} = (\bar{k} \odot \gamma_{K,\ell}) W_{K,\ell}^{(g)}$ per encoder token, for every decoder layer $\ell$ and KV-group $g$, and (under GQA) further replicating them across the $n_h$ Q-heads at attention time. Reverse-QWK instead stores the single, head-agnostic pool $\bar{k} \in \mathbb{R}^{N \times d}$ (Section~\ref{appendix:dimensions_gqa}), where $N$ is the number of tokens in the corpus and $d$ is the embedding dimension. It then pushes the layer/head-specific factors $\gamma_{K,\ell}$ and $W_{K,\ell}^{(g(h))}$ onto the (small) query side via Eq.~\ref{eq:reverse_q}. A single ANN index over $\bar{k}$ therefore serves all decoder layers and Q-heads.

\textbf{Position embeddings:} As in standard encoder-decoder cross-attention, RoPE is applied to the decoder queries $q_\ell$ but not to the encoder representations -- in T5Gemma2 this is enforced by skipping the encoder prefix when applying RoPE to the merged KV stream. Reverse-QWK preserves this: RoPE is applied to $q_\ell$ \emph{before} the transformation in Eq.~\ref{eq:reverse_q}, while $\bar{k}$ remains positionally invariant and can be precomputed once per corpus.

\textbf{Storage footprint and quantization:} Because $\bar{k}$ has no head- or layer-specific axis, its size scales as $N \times d$, independent of $n_h$ and the number of decoder layers. For a 1B-token corpus at $d = 2560$ (the hidden size of our T5Gemma2 4B-4B model) and 8-bit precision, this is $10^9 \times 2560 \approx 2.56$~TB, which fits on a single NVMe SSD; product quantization or further compression could shrink this substantially. Values $V$ are not needed for retrieval and only enter cross-attention at generation time, where they are computed on-demand as $V = W_V \cdot \mathrm{encoder\_kv}$ for the (small) top-$k$ set of selected encoder positions, so they need not be precomputed or stored offline.

\textbf{Compressed cross-attention KV cache for prefill:} Beyond enabling a shared ANN index, $\bar{k}$ can be viewed as a heavily compressed cross-attention KV cache. A standard cross-attention KV cache for $L$ decoder layers stores layer-specific projected $K$ and $V$ for every encoder token, totalling $2L\,n_{\mathrm{kv}} d_h$ scalars per token; Reverse-QWK stores a single $d$-dimensional vector per token, shared across all layers and Q-heads. The resulting compression ratio is $2L\,n_{\mathrm{kv}} d_h / d$, which is exactly $2L$ in the multi-head limit $n_{\mathrm{kv}} d_h = d$ (its theoretical maximum) and somewhat smaller under GQA; for our T5Gemma2 4B-4B model ($L=34$, $n_{\mathrm{kv}} d_h \approx d/2.5$) it is on the order of $\sim\!30\times$. Because $\bar{k}$ already incorporates the encoder forward pass and the only remaining layer-specific work is one query-side multiplication by $W_{K,\ell}$ and $\gamma_{K,\ell}$ (Eq.~\ref{eq:reverse_q}), cross-attention prefill against a cached corpus never re-encodes evidence or recomputes layerwise $K$ -- the prefill cost reduces to a per-layer transformed-query/$\bar{k}$ dot product plus on-demand $V$ for the top-$k$ selected positions.

\section{Retrieval setup}

\subsection{Retrieval Training Details}
\label{appendix:implementation_details}

We initialize from a T5Gemma2 4B-4B checkpoint whose decoder is first
fine-tuned on the CLaRa QA pretraining dataset
\citep{he2026clarabridgingretrievalgeneration}; the configuration of this
warm-start is given in Section~\ref{sec:benchmarks}. It serves three purposes:
aligning the decoder with QA-style generation, which the released checkpoint has
not been post-trained for; adapting generation to chunk-based pre-encoded
representations rather than re-encoded text; and incorporating the Reverse-QWK
reparameterization of cross-attention. The decoder sequence is formatted as
\texttt{Question: \{question\} Answer: \{answer\}}, with the cross-entropy loss taken on the answer tokens only, and an encoder context of 20 context chunks containing both oracle and distractor chunks. We then jointly adapt on the HotPotQA, 2Wiki, MuSiQue, and NQ
training splits using pre-encoded evidence from the shared candidate pool. We train for 15K steps at global batch size 320 with AdamW, learning rate $1\times10^{-5}$ and 1{,}000 warmup steps.

During retrieval training, we optimize only the retrieval-specific parameters: 64 trainable
retrieval token embeddings $\{\rho_i\}$ corresponding to approximately $164K$ parameters, and the learned layer-head aggregation weights $\{\alpha_{l,g}\}$ used
to score chunk vectors corresponding to $272$ parameters. Retrieval training runs for 10K optimization steps with AdamW, learning rate $3 \times 10^{-3}$, 100 warmup steps, and global batch size 256.

The retrieval input in \eqref{eq:x-rho} appends $R$ learnable tokens to $x$, each
producing a query state per decoder layer and head. The retrieval score is computed from the query states at the $R$ retrieval-token positions only, the query states at the input-token positions are not used for scoring. To reduce scoring cost, the $R$
query states are average-pooled into $P \le R$ groups, leaving $P$ vectors per
layer-head that are matched against chunk tokens. $P=R$ keeps one query per retrieval
token; $P=1$ pools all tokens into a single query. The layer-head aggregation with weights $\alpha$ can be applied on either side of the
MaxSim. In the default \emph{sum\,$\to$\,MaxSim} order, the weighted query states are
summed first and MaxSim is evaluated once per pooled query, as in
\eqref{equation:intra_maxsim}. An alternative is \emph{MaxSim\,$\to$\,sum}, in which is evaluated per
layer and head and the scores are combined afterwards, costing one evaluation per
layer-head pair. Summing first is far cheaper and, as
Table~\ref{tab:ablation_chunk_source_at_5_part1} shows, also more accurate.

\subsection{Baseline Details}
\label{appendix:baseline_details}

We compare INTRA against eight retrieval baselines. As sparse lexical baselines we
use TF-IDF \citep{salton1988termweighting} and BM25
\citep{robertson2009bm25}. As dense single-vector baselines we use
BGE-large \citep{xiao2023cpack}, Qwen3-Embedding-0.6B, and
Qwen3-Embedding-4B \citep{zhang2025qwen3embedding}. For BGE and both Qwen
models, we encode the query and each candidate chunk independently and rank all
chunks in the shared pool by cosine similarity. We further evaluate
Qwen3-Embedding-4B with the Jina reranker
\citep{jinaai2024jinarerankerv2}, and a hybrid RAG baseline that combines BM25, TF-IDF, BGE-large, and
Qwen3-Embedding-0.6B rankings using reciprocal rank fusion (RRF)
\citep{cormack2009reciprocal}. We also evaluate a ColBERT-style MaxSim
late-interaction baseline \citep{colbert} that scores query embeddings against
chunk-vector encodings in the T5Gemma2 encoder output space.

\subsection{Initial context selection for retrieval}
\label{sec:context-init} 

In our experiments $\mathcal{S}_0$ is selected with BM25 \cite{robertson2009bm25}. A natural alternative is to select chunks whose encoded representations are closest to the encoded input $x$ in the encoder's own representation space. With $k_x = \mathrm{Enc}(x)$, we define
\begin{equation}\label{eq:context_initial}
    s_i^{\mathrm{enc}} = \mathrm{MaxSim}(k_x, k_i), \qquad
    \mathcal{S}_0^{\mathrm{enc}}
    =
    \left\{ i \in \{1,\ldots,M\} : s_i^{\mathrm{enc}} \text{ is among the top-}n_0 \text{ scores} \right\}.
\end{equation}
This variant keeps the entire pipeline inside one model: initialization, retrieval, and generation all run on the same pretrained encoder-decoder, with no external component of any kind. We find it performs on par with BM25 initialization, but at substantially higher cost to construct $\mathcal{S}_0$. A third option is the empty set $\mathcal{S}_0=\emptyset$, in which case cross-attention is the identity function and $z_\ell=q_\ell$ in Eq.~\ref{eq:xattn-query} (see the ablation study of Table~\ref{tab:ablation_chunk_source_at_5_part1}.)

\subsection{Construction of the candidate set $\mathcal{C}(x)$}
\label{sec:negatives}

For our $\sim$100M-token pool, full-corpus scoring is feasible and we take
$\mathcal{C} = \{1,\dots,M\}$, requiring no negative sampling.

For the full-Wikipedia experiments this is prohibitive, and we build $\mathcal{C}$
per step from the batch. For each query we mine the top-$500$ chunks by BM25 against
its oracle chunks as hard negatives. The per-query pools are merged and deduplicated into a single candidate set shared across the local batch (size $2$ in our experiments), giving $\sim1000$ candidates per step. Only those rows are gathered from the
memory-mapped representation store, so the corpus is never resident in memory.

\section{Additional Results}
\label{appendix:additional_results}

\subsection{Additional retrieval and end-to-end results}

Table~\ref{tab:retrieval_all_oracles_recall}
reports the complete-evidence recall values shown in Figure~\ref{fig:retrieval_recall_all_oracles}. Tables~\ref{tab:e2e_em_at_5_t5gemma} and~\ref{tab:e2e_f1_at_5_t5gemma} present end-to-end question-answering EM and F1 scores, respectively, across various retrieval methods using a fixed T5Gemma2 generator. These tables include the 95\% confidence intervals.

\begin{table*}[t]
    \centering
    \small
    \caption{Complete-evidence recall: the percentage of examples for which all supporting facts are retrieved. INTRA performs best on multi-hop benchmarks (HotPotQA, 2Wiki, MuSiQue) that require evidence assembly. NQ’s single-hop nature minimizes this benefit.
    }
    \label{tab:retrieval_all_oracles_recall}
    \resizebox{\textwidth}{!}{%
    \begin{tabular}{lcccccccccccc}
        \toprule
        & \multicolumn{3}{c}{HotPotQA} & \multicolumn{3}{c}{2Wiki} & \multicolumn{3}{c}{MuSiQue} & \multicolumn{3}{c}{NQ} \\
        \cmidrule(lr){2-4} \cmidrule(lr){5-7} \cmidrule(lr){8-10} \cmidrule(lr){11-13}
        Retrieval method & R@5 & R@10 & R@20 & R@5 & R@10 & R@20 & R@5 & R@10 & R@20 & R@5 & R@10 & R@20 \\
        \midrule
        TF-IDF & 18.2 & 25.9 & 35.5 & 14.0 & 19.3 & 24.7 & 1.5 & 2.4 & 4.8 & 7.5 & 11.3 & 16.2 \\
        BM25 & 32.2 & 41.0 & 48.9 & 17.4 & 23.2 & 28.4 & 3.4 & 5.5 & 7.3 & 14.0 & 21.9 & 32.7 \\
        MaxSim & 36.1 & 46.3 & 54.2 & 16.7 & 22.6 & 27.5 & 5.3 & 8.0 & 11.9 & 20.9 & 29.5 & 39.0 \\
        Hybrid RAG & 48.0 & 61.8 & 71.2 & 29.1 & 36.0 & 40.9 & 6.2 & 11.1 & 17.1 & 22.9 & 35.5 & 49.8 \\
        BGE & 54.8 & 63.5 & 69.6 & 30.9 & 35.9 & 40.1 & 6.5 & 11.2 & 15.7 & 29.6 & 39.0 & 47.5 \\
        Qwen3-Emb-0.6B & 35.9 & 44.2 & 50.4 & 28.0 & 33.3 & 36.8 & 5.7 & 10.7 & 15.5 & 25.6 & 34.4 & 42.0 \\
        Qwen3-Emb-4B & 44.6 & 53.5 & 61.1 & 32.8 & 37.2 & 40.8 & 8.8 & 14.7 & 19.7 & 30.3 & 40.0 & 50.5 \\
        Qwen3-Emb-4B + Jina reranker & 48.8 & 59.6 & 65.4 & 35.4 & 40.3 & 43.5 & 10.1 & 16.6 & 20.6 & \textbf{31.9} & \textbf{42.0} & 50.9 \\
        INTRA & \textbf{70.1} & \textbf{78.7} & \textbf{84.0} & \textbf{50.2} & \textbf{60.0} & \textbf{66.7} & \textbf{16.9} & \textbf{22.9} & \textbf{28.8} & 30.7 & 41.4 & \textbf{51.0} \\
        \bottomrule
    \end{tabular}%
    }
\end{table*}

\begin{table*}[t]
    \centering
    \small
    \caption{End-to-end question-answering EM across retrieval methods with a fixed T5Gemma2 generator. Errors are 95\% CIs.
    }
    \label{tab:e2e_em_at_5_t5gemma}
    \begin{tabular}{lcccc}
        \toprule
        Retrieval method & HotPotQA & 2Wiki & MuSiQue & NQ \\
        \midrule
        TF-IDF & 34.2 $\pm$ 1.0 & 39.0 $\pm$ 0.8 & 5.3 $\pm$ 0.9 & 34.9 $\pm$ 1.2 \\
        BM25 & 40.5 $\pm$ 1.0 & 41.7 $\pm$ 0.9 & 7.7 $\pm$ 1.0 & 43.4 $\pm$ 1.1 \\
        MaxSim & 40.7 $\pm$ 1.1 & 41.6 $\pm$ 0.9 & 10.1 $\pm$ 1.2 & 48.4 $\pm$ 1.2 \\
        Hybrid RAG & 43.4 $\pm$ 1.1 & 46.0 $\pm$ 0.9 & 10.6 $\pm$ 1.2 & 50.5 $\pm$ 1.2 \\
        BGE & 41.9 $\pm$ 1.1 & 46.1 $\pm$ 0.9 & 10.8 $\pm$ 1.2 & 52.2 $\pm$ 1.2 \\
        Qwen3-Emb-0.6B & 37.0 $\pm$ 1.1 & 45.7 $\pm$ 0.9 & 11.1 $\pm$ 1.2 & 36.4 $\pm$ 1.3 \\
        Qwen3-Emb-4B & 40.3 $\pm$ 1.2 & 46.0 $\pm$ 0.9 & 12.7 $\pm$ 1.3 & 54.5 $\pm$ 1.3 \\
        Qwen3-Emb-4B + Reranker & 41.6 $\pm$ 1.2 & 46.8 $\pm$ 1.0 & 13.3 $\pm$ 1.3 & \textbf{55.1 $\pm$ 1.2} \\
        INTRA & \textbf{48.8} $\pm$ \textbf{1.1} & \textbf{52.3} $\pm$ \textbf{0.9} & \textbf{13.9} $\pm$ \textbf{1.3} & 53.7 $\pm$ 1.2 \\
        \bottomrule
    \end{tabular}
\end{table*}

\begin{table*}[t]
    \centering
    \small
    \caption{End-to-end question-answering F1 across retrieval methods with a fixed T5Gemma2 generator. Errors are 95\% CIs.
    }
    \label{tab:e2e_f1_at_5_t5gemma}
    \begin{tabular}{lcccc}
        \toprule
        Retrieval method & HotPotQA & 2Wiki & MuSiQue & NQ \\
        \midrule
        TF-IDF & 44.5 $\pm$ 1.0 & 42.7 $\pm$ 0.7 & 14.6 $\pm$ 1.1 & 42.9 $\pm$ 1.0 \\
        BM25 & 52.0 $\pm$ 0.9 & 45.2 $\pm$ 0.7 & 16.8 $\pm$ 1.2 & 51.8 $\pm$ 1.0 \\
        MaxSim & 52.2 $\pm$ 1.0 & 45.8 $\pm$ 0.8 & 19.8 $\pm$ 1.3 & 57.8 $\pm$ 1.0 \\
        Hybrid RAG & 54.3 $\pm$ 1.0 & 49.9 $\pm$ 0.8 & 20.1 $\pm$ 1.3 & 59.1 $\pm$ 1.1 \\
        BGE & 53.0 $\pm$ 1.0 & 49.6 $\pm$ 0.8 & 19.7 $\pm$ 1.3 & 61.2 $\pm$ 1.1 \\
        Qwen3-Emb-0.6B & 47.7 $\pm$ 1.0 & 49.8 $\pm$ 0.9 & 20.0 $\pm$ 1.4 & 44.6 $\pm$ 1.1 \\
        Qwen3-Emb-4B & 51.2 $\pm$ 1.1 & 50.1 $\pm$ 0.8 & 21.8 $\pm$ 1.4 & 63.7 $\pm$ 1.1 \\
        Qwen3-Emb-4B + Reranker & 53.6 $\pm$ 1.1 & 50.8 $\pm$ 0.8 & 22.5 $\pm$ 1.4 & \textbf{64.2 $\pm$ 1.2} \\
        INTRA & \textbf{60.6} $\pm$ \textbf{1.0} & \textbf{56.4} $\pm$ \textbf{0.8} & \textbf{23.3} $\pm$ \textbf{1.4} & 63.4 $\pm$ 1.1 \\
        \bottomrule
    \end{tabular}
\end{table*}

\begin{figure}[t]
    \centering
    \includegraphics[width=\linewidth]{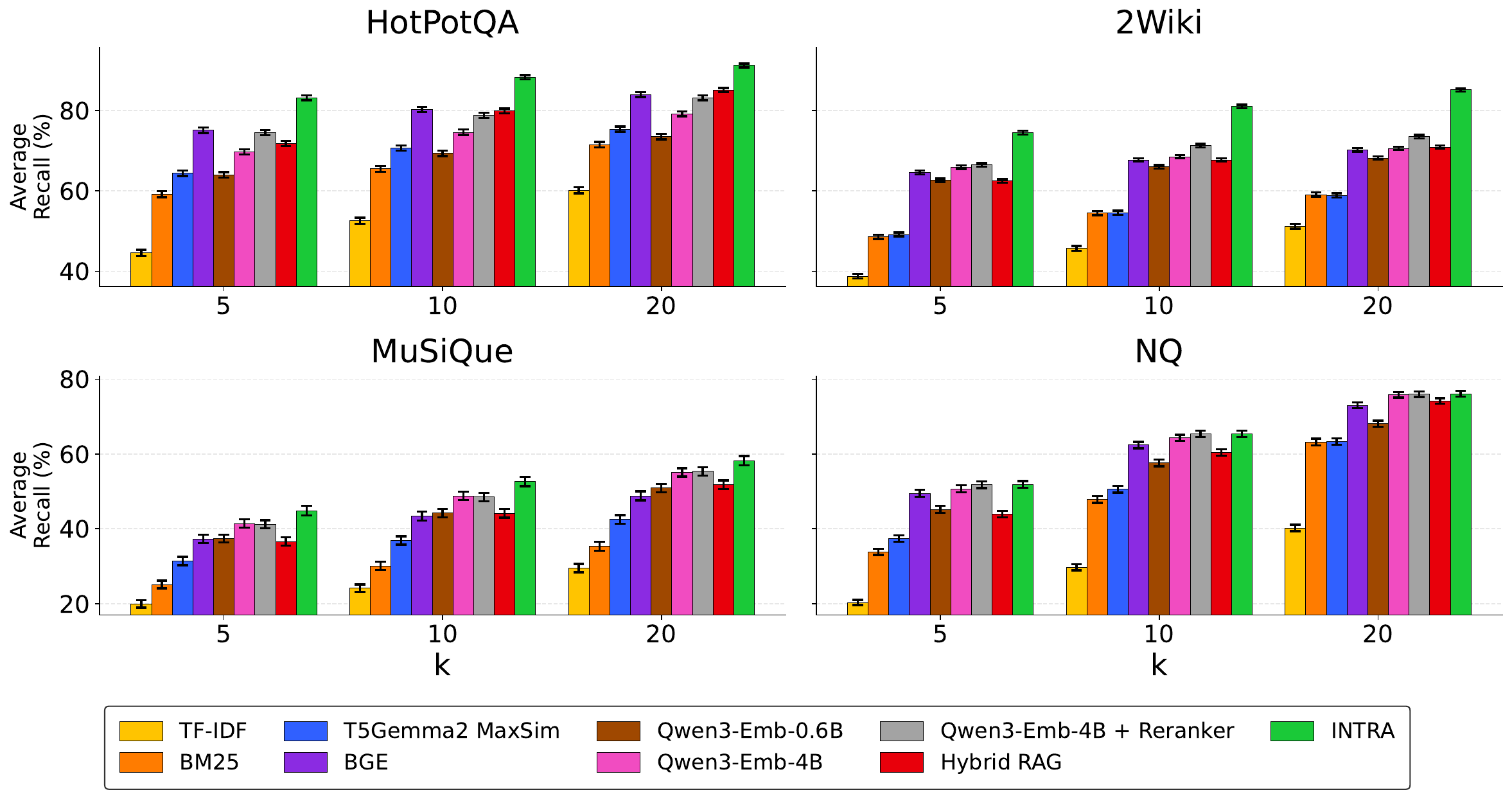}
    \caption{Average recall results on the 100M-token corpus. 
    INTRA performs best on multi-hop benchmarks (HotPotQA, 2Wiki, MuSiQue) that require evidence assembly. 
    NQ’s single-hop nature minimizes this benefit.
    }
    \label{fig:retrieval_recall_average}
\end{figure}

\subsection{Ablation Study}
\label{appendix:ablation_study}

Table~\ref{tab:ablation_chunk_source_at_5_part1} isolates the main design choices in
INTRA's retrieval path. We vary the initial context $\mathcal{S}_0$, the number of
retrieval tokens $R$, the number of groups $P$ into which they are folded, and the
order in which heads and layers are aggregated relative to the MaxSim. Removing
$\mathcal{S}_0$ costs the most recall, and collapsing the retrieval head to a single
token ($P\!=\!R\!=\!1$) is nearly as damaging. Recovering the token budget at $P\!=\!1$
closes roughly half the gap, indicating that both the number of retrieval tokens and
their grouping contribute. Aggregating after the MaxSim rather than before is worse, and reducing $R$ from $64$ to $16$ degrades performance only mildly.

\begin{table}[t]
    \centering
    \small
    \caption{INTRA ablations on HotPotQA and 2Wiki with $k=5$ retrieved chunks, varying the initial set $\mathcal{S}_0$, the retrieval tokens $R$, the pooled queries $P$, and the aggregation order. CE-recall@5 is complete-evidence recall at 5; EM is exact match.
    }
    \label{tab:ablation_chunk_source_at_5_part1}
    \begin{tabular}{llcccc}
        \toprule
        & & \multicolumn{2}{c}{HotPotQA} & \multicolumn{2}{c}{2Wiki} \\
        \cmidrule(lr){3-4} \cmidrule(lr){5-6}
        Design choice & Order & CE-recall@5 & EM & CE-recall@5 & EM \\
        \midrule
        INTRA ($P\!=\!16$, $R\!=\!64$) & Sum $\to$ MaxSim & 70.1 & 48.8 & 50.2 & 52.3 \\
        \midrule
        $\mathcal{S}_0 = \emptyset$ & Sum $\to$ MaxSim & 40.7 \drop{-29.4} &  38.2 \drop{-10.6} & 29.6 \drop{-20.6} & 43.6 \drop{-8.7} \\
        $P\!=\!1$, $R\!=\!1$ & Sum $\to$ MaxSim & 38.7 \drop{-31.4} & 43.0 \drop{-5.8} & 25.0 \drop{-25.2} & 46.8 \drop{-5.5} \\
        $P\!=\!1$, $R\!=\!64$ & Sum $\to$ MaxSim & 53.7 \drop{-16.4} & 45.8 \drop{-3.0} & 39.8 \drop{-10.4} & 50.0 \drop{-2.3} \\
        $P\!=\!1$, $R\!=\!64$ & MaxSim $\to$ Sum & 62.6 \drop{-7.5} & 47.0 \drop{-1.8} & 43.5 \drop{-6.8} & 50.5 \drop{-1.8} \\
        $P\!=\!16$, $R\!=\!16$ & Sum $\to$ MaxSim & 67.7 \drop{-2.4} & 48.5 \drop{-0.3} & 45.7 \drop{-4.6} & 51.6 \drop{-0.7} \\
        \bottomrule
    \end{tabular}
\end{table}

\subsection{Pooled Chunk Length}
\label{appendix:pooled_length}

Table~\ref{tab:ablation_pooled_length} sweeps the pooled chunk length $L_p$ of
Section~\ref{sec:chunk-encoding} from $L_p=1$ to $L_p=128$. At our chunk length of
approximately $130$ tokens, $L_p=128$ is effectively the no-pooling reference, where
$\mathrm{MaxSim}$ is evaluated against every encoder state and \eqref{equation:intra_maxsim}
is exact.

Quality is monotone in $L_p$ over almost the whole range: complete-evidence recall@5
rises with $L_p$ up to $L_p=64$ on both benchmarks and up to $L_p=128$ on HotPotQA, and
EM follows the same ordering. $L_p=7$ is therefore not a quality optimum but a
cost-driven operating point, and pooling to it costs 11.3 points of complete-evidence
recall@5 on HotPotQA and 10.8 points on 2Wiki relative to no pooling.

The cost that grows with $L_p$ is scoring compute and resident GPU memory rather than
disk.
The pooled index is a mean-pooled view of the same per-token states $\bar{k}$ that the decoder cross-attends to during generation, so it does not contain independent data. For our 759K-chunk, 100M-token pool, storing these states requires $512$~GB in \texttt{bf16} at $d=2560$. The pooled copy increases storage by $5\%$ at $L_p=7$ and $49\%$ at $L_p=64$.
At the no-pooling setting the separate copy is redundant altogether, since
retrieval can score the state store itself. What does bind is that the index must be
resident for scoring: at $L_p=7$ it fits on a single device, while the larger settings must
be sharded across devices and nodes.

\begin{table}[t]
    \centering
    \small
    \caption{Pooled chunk length $L_p$ on HotPotQA and 2Wiki with $k=5$ retrieved chunks.
    CE-recall@5 is complete-evidence recall at 5; EM is exact match. Sizes are for the
    759K-chunk, 100M-token pool in \texttt{bf16} at $d=2560$, in decimal GB.
    Total footprint adds the pooled index to the $512$~GB per-token state store that INTRA
    stores for generation regardless of $L_p$; at $L_p=128$ the pooled copy is redundant
    because retrieval scores that store directly. Quality increases monotonically with
    $L_p$, so $L_p=7$ is a scoring-cost operating point rather than a quality optimum.
    This sweep was run with $P=4$.
    }
    \label{tab:ablation_pooled_length}
    \begin{tabular}{lcccccc}
        \toprule
        & \multicolumn{2}{c}{HotPotQA} & \multicolumn{2}{c}{2Wiki} & \multicolumn{2}{c}{Storage} \\
        \cmidrule(lr){2-3} \cmidrule(lr){4-5} \cmidrule(lr){6-7}
        Pooled length & CE-recall@5 & EM & CE-recall@5 & EM & Pooled index & Total \\
        \midrule
        $L_p\!=\!1$ & 48.9 & 43.9 & 26.6 & 45.0 & 3.9 GB & 516 GB \\
        $L_p\!=\!4$ & 57.9 & 46.0 & 37.9 & 48.1 & 15.5 GB & 528 GB \\
        $L_p\!=\!7$ (default) & 59.9 & 46.4 & 40.7 & 49.2 & 27.2 GB & 539 GB \\
        $L_p\!=\!16$ & 65.5 & 47.8 & 48.1 & 50.7 & 62.1 GB & 574 GB \\
        $L_p\!=\!64$ & 67.0 & 48.7 & \textbf{52.0} & \textbf{52.5} & 248 GB & 761 GB \\
        $L_p\!=\!128$ (no pooling) & \textbf{71.2} & \textbf{48.8} & 51.5 & 52.2 & -- & 512 GB \\
        \bottomrule
    \end{tabular}
\end{table}

\subsection{What the Retrieval Scores Attribute, and Which Hop They Find}
\label{appendix:attribution}

Section~\ref{sec:retrieval_results} attributes INTRA's multi-hop advantage to
scoring evidence by the decoder's own attention. We test that directly with a
leave-one-out counterfactual: for every retrieved chunk we remove it from the
generator context and re-score the gold answer, recording the drop in the answer's
log-likelihood and in exact match. This measures how much the answer
depends on a chunk rather than whether it agrees with the annotations. We run it on HotPotQA, with a
20-chunk context, over $7{,}384$ validation questions and $147{,}680$ ablations.

\begin{figure}[t]
    \centering
    \includegraphics[width=\linewidth]{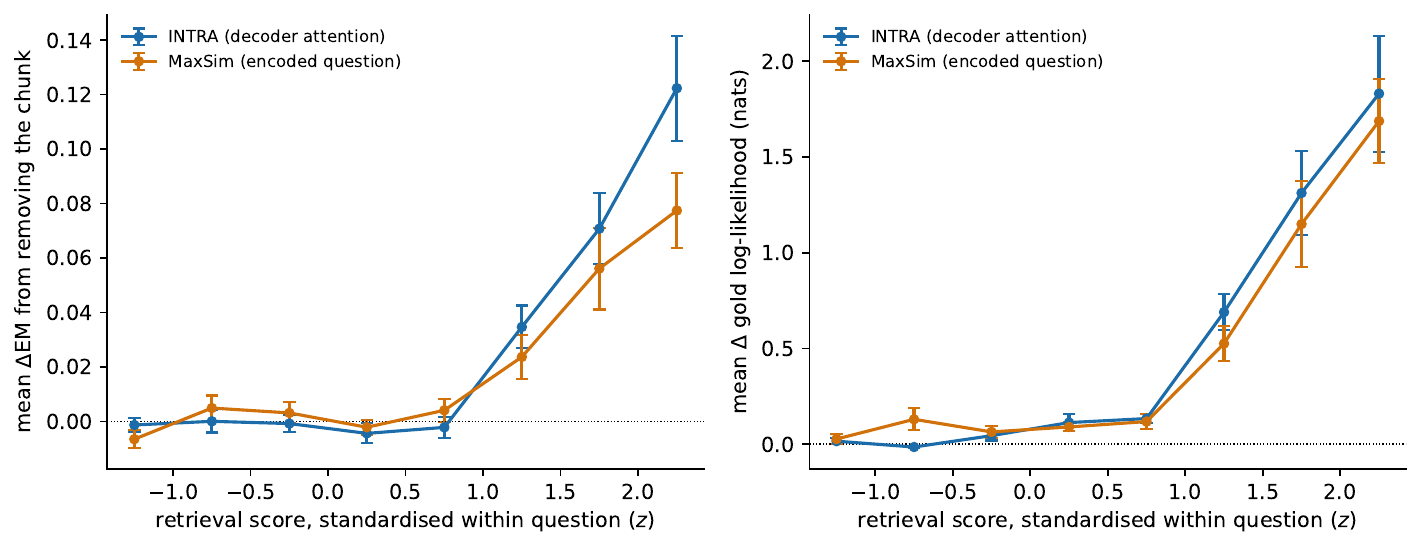}
    \caption{Cost of removing a chunk against its retrieval score, standardized
    within question. \textbf{Left:} exact match. \textbf{Right:} gold answer
    log-likelihood. Error bars are per-question standard errors. Both scorers are
    flat until $z \approx 0.75$ and then rise, with INTRA separating from MaxSim in
    the top bins.}
    \label{fig:loco_em_drop}
\end{figure}

\begin{figure}[t]
    \centering
    \includegraphics[width=\linewidth]{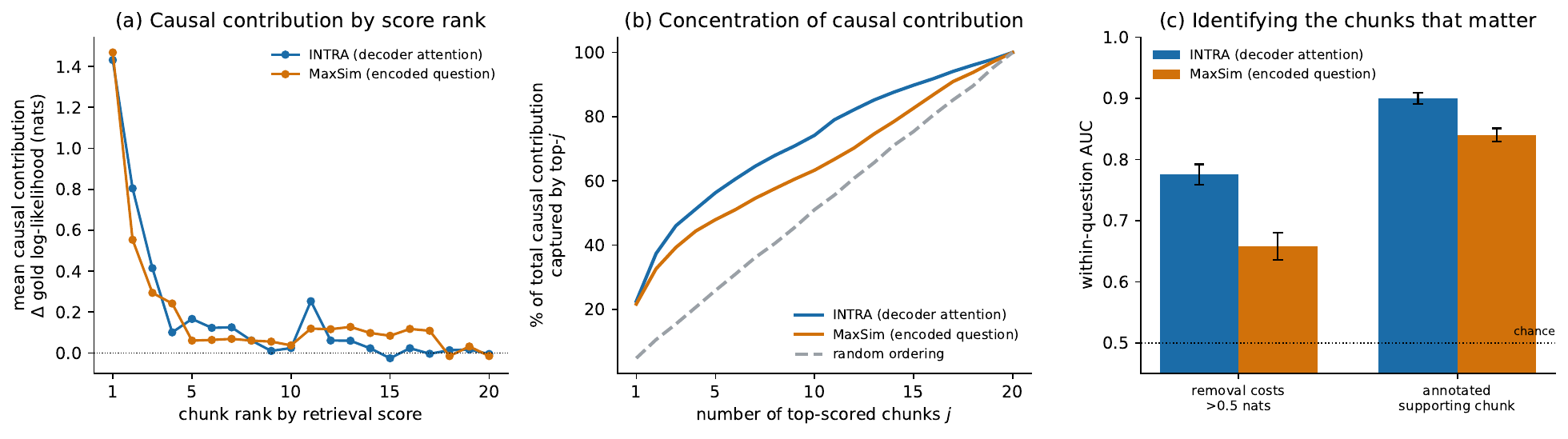}
    \caption{Attribution by rank on the same questions.
    \textbf{(a)} mean log-likelihood drop against rank by retrieval score.
    \textbf{(b)} share of the total drop captured by the top-$j$
    chunks, against a random-ordering reference. \textbf{(c)} within-question AUC
    for identifying chunks whose removal costs more than $0.5$ nats, and for
    identifying annotated supporting chunks, with the $0.5$ chance line.}
    \label{fig:loco_attribution}
\end{figure}

\textbf{The retrieval scores identify chunks the answer depends on.}
Table~\ref{tab:loco_by_rank} and Figure~\ref{fig:loco_em_drop} show that removal cost generally decreases as the retrieval score falls. Removing INTRA’s top-ranked chunk lowers exact match (EM) by $7.2$ percentage points and the gold answer’s log-likelihood by $1.43$ nats. For chunks ranked $6$--$10$, both effects are indistinguishable from zero. This pattern does not arise only because top-ranked chunks are more likely to be annotated as supporting evidence. Even among annotated supporting chunks, removal cost increases as the rank falls: from $+2.06$ nats at rank 1 to $+3.54$ nats at ranks 11--20. In other words, the decoder relies less on supporting chunks that INTRA ranks lower.
Compared with the encoded-question MaxSim baseline, which uses the same representations and generator but a different scoring method, INTRA’s advantage is concentrated among the highest-scored chunks. The overall correlation between retrieval score and EM change is the same for both methods ($r = 0.072$). However, in the top score bin, removing an INTRA chunk causes a $0.045$-point greater EM drop [$+0.010$, $+0.081$]. INTRA also achieves a higher within-question AUC for identifying chunks whose removal lowers log-likelihood by more than $0.5$ nats: $0.78$ versus $0.66$ (Figure~\ref{fig:loco_attribution}c).

\textbf{The two hops are not equally needed.} We divide each two-hop question into
an \emph{answer-bearing} chunk, which contains the answer, and a
\emph{bridge} chunk. We keep the $5{,}288$ questions for which this distinction
is unambiguous and compare INTRA with
NV-ColEmbed-v2\footnote{\url{https://huggingface.co/nvidia/llama-nemotron-colembed-vl-3b-v2}},
a fine-tuned late-interaction retriever that ranks the same pool. Removing the
answer-bearing chunk lowers the gold answer's log-likelihood by $6.05$ nats,
compared with $0.16$ nats for the bridge. At recall@10, INTRA outperforms the
baseline by $12.0$ points on the answer-bearing chunk but trails it by $7.1$
points on the bridge (Table~\ref{tab:loco_per_hop}). The baseline favors the
bridge because the answer-bearing chunk is reached \emph{through} it and shares
little surface form with the question, which favors question-similarity scores.
INTRA therefore allocates more of its retrieval budget to the chunk the answer
depends on most. This trade-off is favorable even when both hops are needed:
INTRA retrieves both at $k=10$ for $64.1\%$ of questions, compared with $54.3\%$
for the baseline.
The scores do not, however, reliably order the selected chunks by how much the
answer depends on them. When both supporting chunks are retrieved, the higher-scored
chunk is the one the answer depends on more in $47.3\%$ of cases [$46.0$, $48.6$],
which is no better than chance. Thus, the useful signal is which chunks enter the
top-$k$, rather than their order within it.

\begin{table}[t]
    \centering
    \small
    \caption{Leave-one-out cost of removing a retrieved chunk from the generator
    context, by INTRA's rank. Scores are standardized within question; ``is
    oracle'' is the share of removed chunks that are annotated supporting chunks.
    Removal cost decays with the score, so the decoder's attention tracks how much
    the answer depends on each chunk.}
    \label{tab:loco_by_rank}
    \begin{tabular}{lcccc}
        \toprule
        Chunk removed, by INTRA rank & Score ($z$) & EM drop & Log-lik. drop & Is oracle \\
        \midrule
        rank 1 & 1.73 & \textbf{7.2\%} & 1.43 nats & 72\% \\
        rank 2 & 1.26 & \textbf{5.7\%} & 0.80 nats & 40\% \\
        rank 3 & 1.00 & 1.3\% & 0.41 nats & 20\% \\
        ranks 4--5 & 0.79 & 0.1\% & 0.13 nats & 8\% \\
        ranks 6--10 & 0.50 & $-0.5\%$ & 0.07 nats & 3\% \\
        ranks 11--20 & $-0.77$ & $-0.1\%$ & 0.04 nats & 1\% \\
        \bottomrule
    \end{tabular}
\end{table}

\begin{table}[t]
    \centering
    \small
    \caption{Per-hop breakdown on the $5{,}288$ HotPotQA questions whose answer
    localizes to exactly one supporting chunk. Answer dependence is the drop in the
    gold answer's log-likelihood when that hop is removed; recall is over all such
    questions, so a hop missing from the pool counts against both retrievers.
    Brackets are $95\%$ bootstrap intervals; ColEmb.\ is NV-ColEmbed-v2. The answer
    depends almost entirely on the answer-bearing hop, which is the hop INTRA gains
    on; it gives up ground on the bridge, which the answer barely uses, and comes out
    ahead when both are required.}
    \label{tab:loco_per_hop}
    \begin{tabular}{lccccccc}
        \toprule
        & & \multicolumn{2}{c}{recall@5} & \multicolumn{2}{c}{recall@10} & \multicolumn{2}{c}{recall@20} \\
        \cmidrule(lr){3-4} \cmidrule(lr){5-6} \cmidrule(lr){7-8}
        Supporting chunk & Answer dependence (nats) & ColEmb. & INTRA & ColEmb. & INTRA & ColEmb. & INTRA \\
        \midrule
        answer-bearing hop & \textbf{+6.05} [+5.75, +6.36] & 55.0 & \textbf{66.1} & 61.9 & \textbf{73.9} & 67.3 & \textbf{76.0} \\
        bridge hop & +0.16 [+0.11, +0.20] & \textbf{87.3} & 78.3 & \textbf{91.1} & 84.0 & \textbf{93.9} & 86.8 \\
        \midrule
        both hops & --- & 44.5 & \textbf{53.3} & 54.3 & \textbf{64.1} & 61.8 & \textbf{67.5} \\
        \bottomrule
    \end{tabular}
\end{table}

\section{Supplementary Efficiency Analysis}
\label{appendix:compute_analysis}
\noindent
Let $N$ denote the
total number of tokens in the evidence pool, let $L_c$ denote the number of
tokens per chunk, let $M \approx N / L_c$ denote the number of chunks, let
$L_q$ denote the query length, let $k = |S(q)|$ denote the number of selected
chunks, and let $L_g$ denote the output length. As in the main text, we express the query-time retrieval cost as $\bigO(\sqrt{M} L_q L_c)$, which is the practical scaling of the MaxSim operator \citep{colbert} when using inverted file (IVF) approximate nearest-neighbor (ANN) indices, such as FAISS or cuVS.

We assume transformer-based encoders and decoders
\cite{NIPS2017_3f5ee243}, so encoding a sequence of length $L$ costs
$\bigO(L^2)$ under dense attention. The goal of this appendix is not to provide
a hardware-faithful latency model, but to make explicit where each family of
methods spends computation. We omit layer, head, and hidden-dimension factors.

\subsection{Full Context Prompting}
Long-context prompting performs no reusable precomputation. The query and all
available evidence are concatenated into one input sequence, so the full cost is
paid at inference time:
\[
\text{Pre-query} = \bigO(1), \qquad
\text{Prefill} = \bigO((N + L_q)^2), \qquad
\text{Generation} = \bigO(L_g (N + L_q + L_g)).
\]
This setting is attractive when the full context must be processed jointly, but
it becomes expensive when relevant evidence is sparse relative to the corpus.

\subsection{Standard RAG}
Standard retrieval-augmented generation preprocesses the corpus into chunks and
encodes them before query time. Under independent fixed-size chunking, this one-time
pre-query work is
\[
\bigO\!\left(\frac{N}{L_c} \cdot L_c^2\right) = \bigO(NL_c).
\]
At inference time, RAG first pays retrieval cost
$\bigO(\sqrt{M} L_q L_c)$ and then re-encodes the retrieved text together with
the query:
\[
\text{Prefill} = \bigO((L_q + kL_c)^2), \qquad
\text{Generation} = \bigO(L_g (L_q + kL_c + L_g)).
\]
Relative to long-context prompting, this reduces the amount of evidence
re-encoded per query, but still requires the generator to process the retrieved
raw text from scratch. The main difference from INTRA is the quadratic encoder-side
re-encoding term.

\subsection{INTRA}
INTRA shares the same offline chunk-encoding term as standard RAG,
\[
\text{Pre-query} = \bigO(NL_c),
\]
but changes what happens after retrieval. The model retrieves reusable chunk
representations rather than raw passages, so query-time work becomes
\[
\text{Retrieval} = \bigO(\sqrt{M} L_q L_c), \qquad
\text{Prefill} = \bigO(L_q(L_q + kL_c)), \qquad
\text{Generation} = \bigO(L_g (L_q + kL_c + L_g)).
\]
The key distinction is that evidence is encoded once offline and then reused at
query time, which shifts work away from repeated evidence prefilling and toward
reusable memory construction.

\subsection{Additional Timing Results and Benchmark Details}
\label{appendix:additional_time_results}
\label{appendix:benchmark_details}

The efficiency benchmark (Figure~\ref{fig:bench_ttft_k} in the main text) starts
from $L_q=L_c=L_g=128$, $N=65{,}536$ and $k=20$, sweeps one of these per figure,
excludes retrieval time, and gives every system the same top-$k$ chunks. It
therefore measures only the generator-side cost after evidence has been selected. Full-context prompting is the reference point,
incurring $\bigO((L_q+N)^2)$ prefill by processing the query with the whole
evidence pool; it is independent of both $k$ and $L_c$, so it appears throughout
as a flat line at 8.21\,s of prefill, 8.2\,tok/s of decode and 63.6\,GB of the
80\,GB device. Standard RAG pre-fills over
the query and $k$ retrieved chunks, $\bigO((L_q+kL_c)^2)$, whereas INTRA reuses
stored chunk states and pays only $\bigO(L_q^2)$ before the first token.

\begin{figure}
    \centering
    \includegraphics[width=\linewidth]{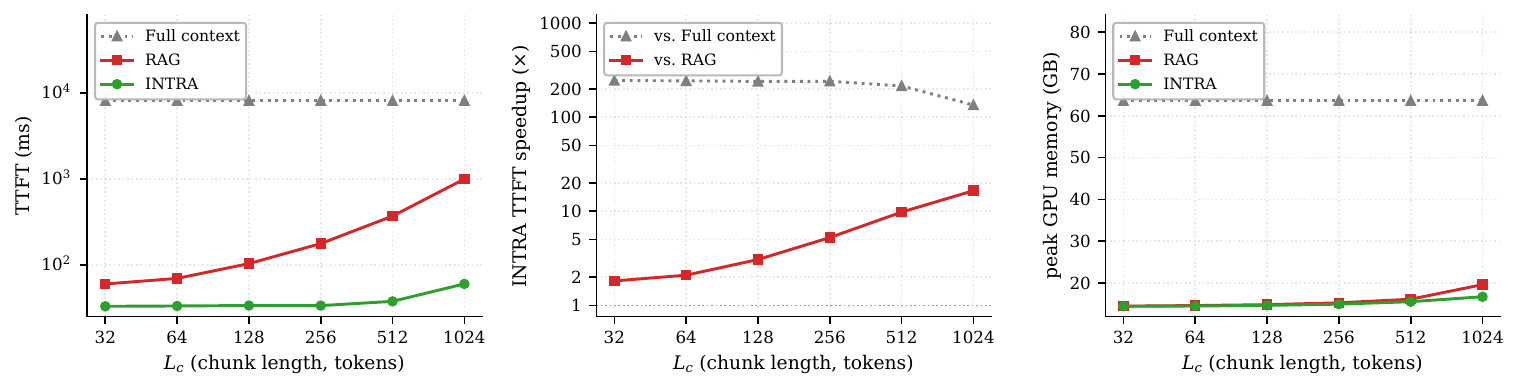}
    \caption{Prefill cost against chunk length $L_c$ at $k=20$, at T5Gemma2
    4B-4B on one H100, in the layout of Figure~\ref{fig:bench_ttft_k}.
    \textbf{Left:} time to first token, where re-encoding grows with $kL_c$ while
    INTRA stays flat until the memory itself becomes large.
    \textbf{Middle:} INTRA's TTFT speedup. \textbf{Right:} peak GPU memory, of
    which 14.2~GB is model weights.}
    \label{fig:bench_ttft_L_c}
\end{figure}

The measured curves follow the asymptotic comparison. As $k$ increases from 1 to
500, INTRA TTFT grows from 33.8\,ms to 156\,ms, while standard RAG grows from
61.5\,ms to 7.81\,s, so INTRA's speedup over re-encoding widens from $1.8\times$
to $50\times$ and its advantage over full-context prompting narrows from
$243\times$ to $53\times$. Figure~\ref{fig:bench_ttft_L_c} shows the same trend as
chunk length increases at a fixed $k=20$, with re-encoding reaching 1.01\,s at
$L_c=1024$ against 60.8\,ms for INTRA. All methods use the same Reverse-QWK
decoder, which lets normalized encoder states serve as reusable cross-attention
memory without precomputing separate layer-specific projected key/value states.

\begin{figure}
    \centering
    \includegraphics[width=\linewidth]{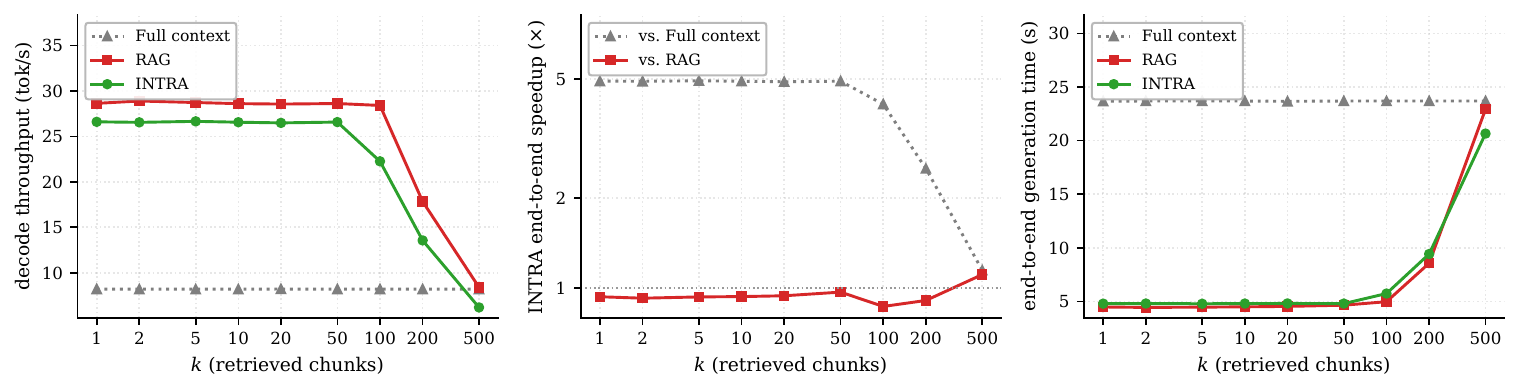}
    \caption{Generation cost against the number of retrieved chunks $k$, at
    T5Gemma2 4B-4B on one H100. \textbf{Left:} decode throughput, measured after
    the first token. \textbf{Middle:} INTRA's end-to-end speedup, prefill plus
    128 generated tokens. \textbf{Right:} the end-to-end time that ratio is taken
    from.}
    \label{fig:bench_decode_k}
\end{figure}

\textbf{Decoding.} Figure~\ref{fig:bench_decode_k} reports tokens per second after
the first token, where the gap is far smaller than in prefill: 26.5\,tok/s for
INTRA at $k=20$ against 28.5 for re-encoded RAG, since scoring the shared
$d$-dimensional pool and projecting $V$ on demand is slightly more work per token
than attending to re-encoded states. Over 128 generated tokens this outweighs the
prefill saving at moderate $k$, so end-to-end INTRA is $0.94\times$ re-encoded RAG
at $k=20$ and overtakes it only at $k=500$ ($1.11\times$), while remaining
$4.9\times$ faster than full-context prompting throughout. INTRA's efficiency
claim is therefore about prefill and storage rather than decoding.

\subsection{Comparison Against a Fully Cached Baseline}
\label{appendix:cached_baseline}

Re-encoding is not the only way to serve the selected chunks: their
cross-attention states can also be precomputed and cached.
Table~\ref{tab:ttft_cached} adds this baseline in its strongest form, with every
decoder layer's $K$ and $V$ stored so that prefill performs no encoding and no
projections. We report it separately from the figures above because it is not a
system we compare against on quality, and because the comparison turns on data
movement rather than on the compute those figures measure. At corpus scale neither
INTRA's pool nor the cached keys and values fit in HBM, so both fetch the states of
their selected chunks from local storage per query, and the measurements include
that fetch. Caching removes essentially all of the re-encoding gap, and below
$k \approx 8$ it is marginally faster than INTRA; over that range,
Figure~\ref{fig:bench_ttft_k} should be read as an advantage over re-encoding
rather than over caching. It sustains that latency only by moving $27.2\times$ more
state per query, however, so it crosses over at $k \approx 8$ and is $1.5\times$
slower by $k=100$: INTRA's latency advantage at practical $k$ is a consequence of
its storage advantage. Where that storage gap is small the balance shifts back into
a memory-for-latency trade-off, which is the regime of very small $k$ and of
architectures whose per-token cache is already thin, such as multi-query or
multi-head latent attention \citep{deepseekai2024deepseekv3technicalreport}.

\begin{table}[t]
    \centering
    \small
    \caption{Time to first token and state fetched per query, against a cached baseline that precomputes every decoder layer's cross-attention $K$ and $V$ so that prefill performs no encoding and no projections. Measured at the paper's configuration (T5Gemma2 4B-4B, one H100), including the per-query fetch of selected chunk states from local NVMe, at $L_c=128$ and 8-bit precision. Caching matches INTRA below $k \approx 8$, but holds that latency only by moving $27.2\times$ more state per query.}
    \label{tab:ttft_cached}
    \begin{tabular}{lccccc}
        \toprule
        Retrieved chunks $k$ & 5 & 10 & 20 & 50 & 100 \\
        \midrule
        \multicolumn{6}{l}{\emph{Time to first token} (ms)} \\
        Standard RAG (re-encode) & 60.9 & 69.3 & 103.3 & 226.9 & 518.5 \\
        Standard RAG + cached $K$, $V$ & \textbf{32.4} & 34.6 & 38.4 & 50.3 & 69.8 \\
        INTRA & 33.8 & \textbf{33.9} & \textbf{34.4} & \textbf{35.1} & \textbf{45.5} \\
        \midrule
        \multicolumn{6}{l}{\emph{State fetched per query} (MB)} \\
        Standard RAG + cached $K$, $V$ & 44.6 & 89.1 & 178 & 446 & 891 \\
        INTRA & \textbf{1.6} & \textbf{3.3} & \textbf{6.6} & \textbf{16.4} & \textbf{32.8} \\
        \bottomrule
    \end{tabular}
\end{table}
\subsection{Compute Resources}
\label{appendix:compute_resources}

Training was run on NVIDIA H100 GPUs with 80GB of memory per GPU. The largest
training configuration used up to 160 GPUs for the QA pretraining stage and for
each retrieval/generation model version. The longest training run was the QA
pretraining stage, which took approximately 12 hours, corresponding to about
2{,}000 GPU-hours. The benchmark-specific retrieval and generation training runs
were much shorter, each accounting for less than 200 GPU-hours. Our training
code uses a modified implementation based on TorchTitan \citep{torchtitan}.

\section{Dataset and Pool Details}
\label{appendix:clara_dataset_details}

We use a deduplicated end-to-end QA dataset derived from the union of the
per-benchmark datasets constructed in CLaRa
\citep{he2026clarabridgingretrievalgeneration}, with one training split and one
evaluation split for each of HotPotQA, 2WikiMultihopQA, MuSiQue, and Natural
Questions.

Following the setup of CLaRa, we build one shared retrieval candidate pool for all four benchmarks under a
fixed budget of approximately 100M tokens. We start from the
benchmark-specific chunk sets provided for HotPotQA, 2Wiki, MuSiQue, and NQ,
merge them into one pool, and de-duplicate chunks globally. We ensure coverage of
all oracle chunks referenced by the QA examples and add uniformly sampled
non-oracle chunks without replacement until the budget is reached.

\begin{table}[h]
    \centering
    \small
    \caption{Split statistics for the deduplicated end-to-end QA dataset used in our experiments.}
    \label{tab:clara_100m_split_stats}
    \begin{tabular}{lrr}
        \toprule
        Split & Examples & Doc refs \\
        \midrule
        2Wiki train & 167,454 & 404,170 \\
        HotPotQA train & 90,185 & 180,370 \\
        MuSiQue train & 277,577 & 631,153 \\
        NQ train & 53,301 & 276,872 \\
        2Wiki eval & 12,576 & 30,654 \\
        HotPotQA eval & 7,384 & 14,768 \\
        MuSiQue eval & 2,417 & 6,404 \\
        NQ eval & 6,489 & 42,182 \\
        \bottomrule
    \end{tabular}
\end{table}

Across all splits, the saved dataset contains 617K QA examples and
1.6M document references. These references cover 514,999 unique oracle
chunks. The global context pool contains 759K chunks in total: all 514,999
oracle chunks plus 243,501 non-oracle chunks sampled to reach the 100M-token
pool budget. The pool contains 69,734,081 whitespace-delimited words
(91.94 words per chunk on average, 106 median). Note that while 243K
chunks are non-oracle for every example in the dataset, from the perspective
of any single example all other $\sim$759K chunks in the pool, including
those that serve as oracles for other examples, act as distractors.

\section{Broader Societal Impact}
\label{appendix:broader_impact}

INTRA is a retrieval-and-generation method rather than a deployed application,
but it may affect systems that answer questions over large knowledge bases. On
the positive side, reusing model-native evidence representations could reduce
the repeated computation needed to serve retrieval-augmented QA over stable
corpora, potentially lowering cost and making high-quality evidence access more
practical. On the negative side, stronger retrieval-conditioned generation can
also make incorrect, biased, private, or otherwise sensitive evidence easier to
surface and present fluently if the underlying corpus or deployment safeguards
are inadequate. As with other RAG systems, deployment should therefore include
corpus governance, access controls, provenance tracking, and application-level
evaluation for hallucination, privacy, and fairness risks.

\newpage

\end{document}